\documentclass[10pt,twocolumn,letterpaper]{article}

\usepackage[pagenumbers]{cvpr} 

\usepackage{multirow}
\usepackage{colortbl}
\definecolor{cvprblue}{rgb}{0.21,0.49,0.74}
\usepackage[pagebackref,breaklinks,colorlinks,allcolors=cvprblue]{hyperref}
\usepackage[normalem]{ulem} 

\def\paperID{*****} 
\def\confName{CVPR}
\def\confYear{2026}

\usepackage{algorithm}
\usepackage{xcolor}
\usepackage{caption}

\usepackage{listings}
\usepackage{xcolor}

\definecolor{codebg}{RGB}{245,245,245}
\definecolor{pycomment}{HTML}{3D7B7B}
\definecolor{pykeyword}{HTML}{008000}
\definecolor{pyfunction}{HTML}{0000FF}
\definecolor{pystring}{HTML}{BA2121}
\definecolor{pynumber}{HTML}{666666}

\lstdefinestyle{pythonminted}{
    language=Python,
    backgroundcolor=\color{codebg},
    basicstyle=\small\ttfamily,
    keywordstyle=\color{pykeyword}\bfseries,
    commentstyle=\color{pycomment}\itshape,
    stringstyle=\color{pystring},
    showstringspaces=false,
    breaklines=true,
    columns=fullflexible,
    keepspaces=true,
    morekeywords=[2]{DISReg,DiffEnc,DiffPred,SIGReg, mean},
    keywordstyle=[2]\color{pyfunction},
}

\newcommand{\modelname}{MotionJEPA\xspace}
\newcommand{\componentname}{DISReg\xspace}   

\definecolor{codebg}{HTML}{F4F8FA}

\title{\modelname: Preventing Temporal Feature Collapse by Capturing Visual Changes in Latent Space}

\author{
  \textmd{
    \large Markus Karmann\textsuperscript{1,*} \qquad
    Shile Li\textsuperscript{2,*} \qquad
    Christian Internò\textsuperscript{3,1} \qquad
    Bruno Andreis\textsuperscript{1,4} \qquad
  } \\[0.5ex]
  \textmd{
    \large David Klindt\textsuperscript{5} \qquad
    Randall Balestriero\textsuperscript{6,7} \qquad
    Jindong Gu\textsuperscript{1} \qquad
    Philip Torr\textsuperscript{1,4} \qquad
  } \\[0.5ex]
  \textmd{
    \large Qi Zhang\textsuperscript{8} \qquad
    Peng-Tao Jiang\textsuperscript{8} \qquad
    Hao Zhang\textsuperscript{8} \qquad
    Bo Li\textsuperscript{8} \qquad
    Onay Urfalioglu\textsuperscript{2}
  } \\[3ex]
  \textsuperscript{1}University of Oxford \qquad
  \textsuperscript{2}vivo Tech Research GmbH \qquad
  \textsuperscript{3}Bielefeld University \qquad
  \textsuperscript{4}Slater Labs \\[0.5ex]
  \textsuperscript{5}Cold Spring Harbor Laboratory \qquad
  \textsuperscript{6}Brown University \qquad
  \textsuperscript{7}AMI Labs \\[0.5ex]
  \textsuperscript{8}vivo BlueImage Lab, vivo Mobile Communication Co., Ltd., China \\[1.5ex]
  \textsuperscript{*}Equal contribution
}

\begin{document}
\maketitle
\begin{abstract}
Joint Embedding Predictive Architectures (JEPAs) are a promising paradigm for learning task-agnostic latent world models without visual reconstruction.
However, standard JEPA training exhibits a strong inductive bias towards slow features, causing feature suppression and the collapse of latent representation.
While inverse dynamics provides temporal anti-collapse, it relies on action labels and offers little incentive to embed general, unlabeled dynamics.
We introduce \textbf{D}ifference \textbf{I}mage and \textbf{S}ingle image embedding \textbf{Reg}ularization (\componentname), a novel regularizer that builds on an inverse-dynamics-style module that predicts temporal difference image embeddings without any pixel reconstruction loss, encouraging balanced static and dynamic feature learning.
\componentname consists of a static term that shapes the distribution of the image embedding and encourages slow features, and a dynamic term, which, unlike direct regularization on the embedding, imposes no constraint on the image embedding's shape or distribution and instead only incentivizes that dynamic features be present.
By integrating this regularizer into a standard JEPA, we establish our new architecture, \modelname.
Latent probing demonstrates that \modelname produces more complete representations than other methods, and our trajectory analysis shows it maintains geometrically simple latent embeddings with low curvature.
We further show that \modelname improves downstream planning success under static-background distractors across four environments.
Code available: \url{https://github.com/mkarmann/motion-jepa}.
\end{abstract}

\begin{figure}[h]
    \centering
    \includegraphics[width=0.94\linewidth]{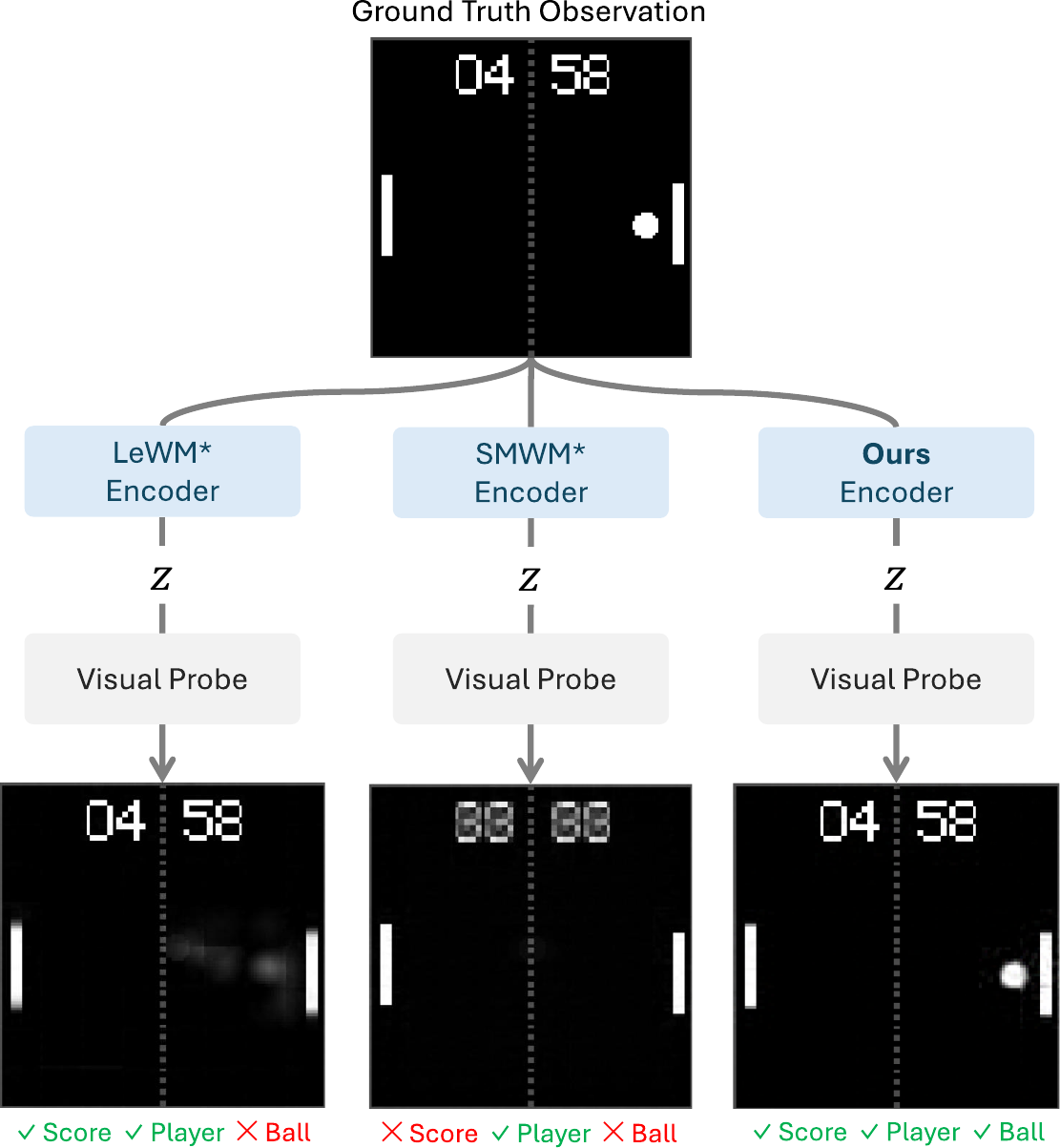}
    \caption{\textbf{Feature Suppression}. We compare three JEPA variants that differ in collapse-prevention mechanisms: LeWM~\cite{LeWorldModel} (SIGReg), SMWM~\cite{SMWM} (Inverse Dynamics Module, IDM), and our \modelname, which uses \componentname. The Pong environment contains three features: a fast-moving ball, two bats, and a slow-changing scoreboard. When trained from scratch, visual probes of the image embeddings show that SIGReg encodes the slowly changing scoreboard and bats but suppresses the ball. IDM encodes only the action-controlled bats. \componentname preserves all three.}
    \label{fig:thumbnail}
\end{figure}

\section{Introduction}
\label{sec:intro}
Joint Embedding Predictive Architectures (JEPAs)~\cite{APathTowardsAutonomousMachineIntelligence} have emerged as a powerful paradigm for representation learning by predicting future states directly within an abstract embedding space.
However, a fundamental question in training JEPA architectures lies in determining exactly which features the model should capture.
Previous works such as ~\cite{CuriosityDrivenExloration} suggest encoding features that are either controlled by the agent or can affect the agent, ignoring uncontrollable background features.
Yet the optimal feature hierarchy is inherently problem- and task-dependent.
Short-time planning operates on a short timescale, predicting fast-changing dynamics, whereas long-time planning operates on larger timescales with slower-changing features~\cite{APathTowardsAutonomousMachineIntelligence}.
Although the primary objective of JEPA frameworks is to learn task-agnostic world models, standard formulations exhibit a well-documented inductive bias toward encoding slow features~\cite{JEPAsFocusOnSlowFeatures,TheObsessedEncoder}.
In JEPA architectures, this trade-off is mainly controlled by the collapse prevention mechanism~\cite{JEPAsFocusOnSlowFeatures}.
We illustrate this in \Cref{fig:thumbnail}, comparing SIGReg-based LeWorldModel~\cite{LeWorldModel}, inverse dynamics-based SMWM~\cite{SMWM}, and our \componentname-based \modelname.
The \textbf{D}ifference \textbf{I}mage and \textbf{S}ingle image embedding \textbf{Reg}ularization (\componentname) is a novel regularizer that encourages a complementary and balanced embedding of both static and dynamic features.
\componentname is made of two terms.
The static term shapes the distribution of the single-image embedding, encouraging the slow features that standard JEPA training already favors.
The dynamic term instead predicts the embedding of the temporal difference image from consecutive image embeddings.
This dynamic term does not constrain the shape or distribution of the image embedding and only incentivizes that features that change between frames are present, which we show results in geometrically simple latent trajectories with low curvature in \Cref{tab:path_straightness}.\\
\\
\noindent Our main contributions are as follows:
\begin{itemize}
    \item \textbf{\componentname}: A novel regularization term that incentivizes a balanced embedding of static and dynamic features by utilizing embeddings of the temporal difference images.
    \item \textbf{\modelname}: A novel JEPA framework that utilizes \componentname, demonstrating better downstream performance in the presence of static distractors and robustness to feature collapse.
    \item A detailed collapse and latent trajectory analysis on a novel collection of three synthetic game environments (Pong, Dino, and Golf) with clearly separable, well-defined feature groups.
    \item Downstream control evaluation on experiments demonstrating the strengths of \modelname in the presence of static distractors.
\end{itemize}


\begin{figure}[t!]
  \centering
  \includegraphics[width=1\linewidth]{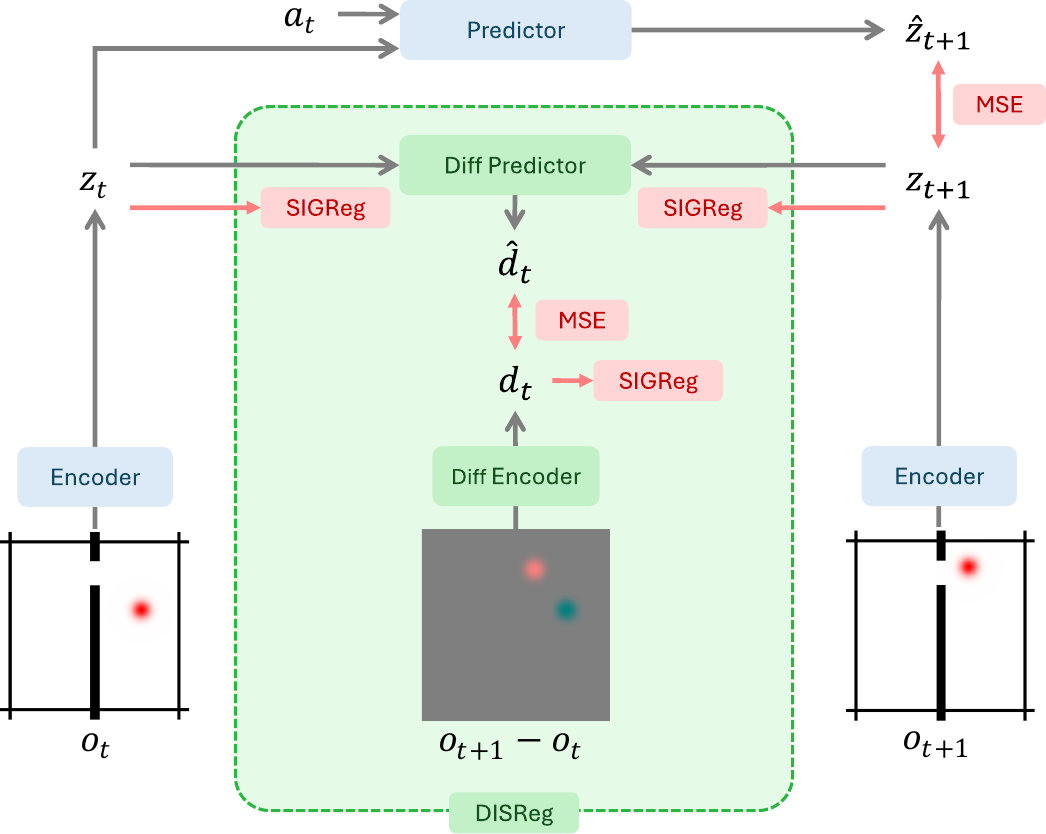}
  \caption{\textbf{\modelname Overview}. Our novel regularizer \componentname utilizes a temporal difference image embedding $d_t$ as visually grounded collapse prevention for temporally changing features. The core JEPA encoder and predictor architecture concept is illustrated in blue, and our \componentname in green. The JEPA forward prediction loss, as well as \componentname losses are highlighted in red.}
  \label{fig:framework_overview}
\end{figure}

\section{Related Work}
\label{sec:related_work}

\paragraph{JEPA-style Latent World Models.}
The Joint Embedding Predictive Architecture (JEPA) is an architectural design that aims to learn latent representations without reconstructing the observation.
As latent world models, these models learn to predict the dynamics of a system in latent space.
The V-JEPA models~\cite{VJEPA, VJEPA2, VJEPA21} utilize a two-stage training paradigm by first pre-training on unlabeled video and then fine-tuning for the desired task.
Recent work has further explored latent predictive models for control and planning, including LeWorldModel~(LeWM)~\cite{LeWorldModel}, which learns action-conditioned dynamics directly in representation space, and DINO-WM~\cite{DINOWM}, which builds a world model on top of pretrained visual features for downstream planning tasks with Model Predictive Control (MPC).
Temporal Difference in Vision (TDV)~\cite{TDV} does not require action labels and instead uses the pixel difference between consecutive frames to condition latent prediction and predict the corresponding change in representation space.
Our work similarly exploits visually derived temporal change, but uses difference images as an auxiliary representation-learning objective to encourage the latent space to preserve changing scene information without requiring action labels.
 
\paragraph{Collapse Prevention in JEPA.}
Without a reconstruction loss, the main challenge of JEPAs is to avoid convergence to a trivial solution where all inputs map to a constant latent representation.
Contrastive methods use negative samples to explicitly push embeddings apart~\cite{SimCLR}.
Non-contrastive methods utilize stop-gradient teachers~\cite{IJEPA, VJEPA, VJEPA2, VJEPA21, TDV}.
Other approaches avoid negative pairs entirely, such as VICReg, which prevents informational collapse by applying explicit variance, invariance, and covariance regularization directly to the embeddings~\cite{VICReg}.
More recent methods avoid computing heavy covariance matrices by shaping the latent space directly into an isotropic Gaussian distribution.
Specifically, LeJEPA~\cite{LeJEPA} introduces SIGReg as its sole collapse-prevention loss, using the Epps-Pulley test~\cite{EppsPulley} rather than the Sliced Wasserstein Distance (SWD)~\cite{SWD} to sketch the distribution.
However, avoiding complete representational collapse does not ensure that the learned representation preserves all task-relevant factors of variation.

\paragraph{Feature Suppression and Slow-Feature Bias.}
Beyond complete collapse, self-supervised models may exhibit \emph{feature suppression}, retaining only a subset of relevant information despite globally diverse representations~\cite{FeatureSuppression,ContrastiveShortcuts}.
In temporal predictive learning, this can appear as a preference for slowly varying features.
While JEPA performs on par with or better than reconstruction-based methods in settings where noise changes every time step, JEPA-style models fail when fixed noise or features are present.
\citet{JEPAsFocusOnSlowFeatures} show that JEPAs can encode fixed trajectory-specific distractors while neglecting faster foreground dynamics, a failure later reproduced across different noise timescales and predictive architectures~\cite{ContrastInductiveBias}.
Similar sensitivity to background changes has been observed in JEPA-based planning~\cite{InvariantJEPAPlanning}, while recent theory further connects LeJEPA's temporal alignment objective to Slow Feature Analysis~\cite{WhenLeJEPALearnsWorldModel}.
LeNEPA~\cite{LeNEPA} applies SIGReg along the time dimension instead of the batch dimension to encourage temporal diversity in the learned representation.

Inverse dynamics provides another way to counteract this bias by forcing the representation to preserve information about action-relevant changes.
Sensorimotor World Models (SMWM)~\cite{SMWM} introduces an inverse model which predicts the action of two consecutive frames, given only the image embeddings.
Experiments show that this additional module prevents collapse and enables the model to learn actual controllable degrees of freedom.
Similarly, Delta-JEPA~\cite{DeltaJEPA} also relies purely on the action prediction for collapse prevention, but provides the difference of two consecutive embeddings as input to the action decoder.
A core limitation of inverse action prediction is that action labels are required.
Additionally, these methods may be less suitable in scenarios where a given action causes a reaction in the observation with a time delay.

These observations motivate our evaluation setting, where we introduce static background distractors into control environments to explicitly test whether the learned representation preserves changing, task-relevant scene information in the presence of strong slow features.
Motivated by the same limitation, our  method introduces a visually grounded temporal-difference objective that explicitly encourages the latent representation to preserve changing scene information without requiring action supervision.

\section{Method}
\label{sec:method}

\paragraph{Problem formulation.}
We are given an offline dataset of trajectories $\mathcal{D} = \{(o_1, a_1, \dots, o_T)\}$ with visual observations $o_t \in \mathbb{R}^{3 \times H_\text{obs} \times W_\text{obs}}$ and continuous action labels $a_t \in \mathbb{R}^{D_a}$.
The trajectories are not required to follow a task-specific policy and can be collected from any unknown or arbitrary behavior policy.
The goal is to train a JEPA-style latent world model consisting of an encoder $\text{Enc}_\theta: \mathbb{R}^{3 \times H_\text{obs} \times W_\text{obs}} \rightarrow \mathbb{R}^{D_z}$ that maps individual observations $o_t$ to a latent state $z_t \in \mathbb{R}^{D_z}$, and a forward predictor $\text{Pred}_\phi$ that estimates the deterministic state transition $\hat{z}_{t+1} = \text{Pred}_\phi(z_{t-H+1:t}, a_{t-H+1:t})$ where $H$ is the size of the history context window.
In the JEPA setting, this training process does not involve any reconstruction loss on the visual observations $o_t$, resulting in the primary challenge of preventing $z_t$ from collapsing to trivial solutions~\cite{APathTowardsAutonomousMachineIntelligence}.

\paragraph{Collapse prevention with \componentname.}
As illustrated in \cref{fig:framework_overview}, our proposed \componentname introduces two new components, an encoder for encoding the temporal difference image $\text{DiffEnc}_{\alpha}: \mathbb{R}^{3 \times H_\text{obs} \times W_\text{obs}} \rightarrow \mathbb{R}^{D_d}$ and an inverse dynamics module-inspired predictor $\text{DiffPred}_{\beta}: \mathbb{R}^{D_z} \times \mathbb{R}^{D_z} \rightarrow \mathbb{R}^{D_d}$.
\begin{align}
    d_t &= \text{DiffEnc}_{\alpha}(o_{t+1} - o_t)\\
    \hat{d}_{t} &= \text{DiffPred}_{\beta}(z_t, z_{t+1})
\end{align}
$\text{DiffPred}_\beta$ learns to predict the difference latent $d_t$ of the difference image given the corresponding state embeddings $z_t$ and $z_{t+1}$.
This design has the advantage that we do not require action or state labels as targets, therefore introducing no bias towards labeled features.

\begin{algorithm}[!t]
\caption{\textbf{Pseudocode of DISReg}. Inputs are an observations tensor $o \in \mathbb{R}^{N \times T \times 3 \times H_\text{obs} \times W_\text{obs}}$ with the per-image embeddings tensor $z \in \mathbb{R}^{N \times T \times D_z}$.} \label{alg:disreg}
\begin{lstlisting}[style=pythonminted]
def DISReg(o, z):
    # 1. Compute temp. difference images
    o_current = o[:, :-1]
    o_next = o[:, 1:]
    o_diff = o_next - o_current

    # 2. Embed difference images
    d = DiffEnc(o_diff)

    # 3. Predict dynamics from latents
    z_current = z[:, :-1]
    z_next = z[:, 1:]
    d_pred = DiffPred(z_current, z_next)
    
    # 4. Estimate losses
    L_z = SIGReg(z)
    L_d = SIGReg(d)
    L_pred = (d - d_pred).pow(2).mean()
    
    # 5. Aggregate losses
    return 0.25*L_z + 2*L_d + 0.5*L_pred
\end{lstlisting}
\end{algorithm}

\paragraph{Training losses.}
In contrast to a conventional JEPA design, our architecture has two latent representations $z$ and $d$ that are prone to collapse.
We choose SIGReg~\cite{LeJEPA} for collapse prevention, but other distribution regularization losses~\cite{VISReg, SWD} can also be utilized.
\begin{align}
    \mathcal{L}_z = \text{SIGReg}(z) \;\;\;\;\;\;\;\;\; \mathcal{L}_d = \text{SIGReg}(d)
\end{align}
We establish the prediction loss between the temporal difference image embedding $d_t$ and the respective predictor's embedding $\hat{d}_t$ and aggregate them in $\mathcal{L}_{\text{\componentname}}$.
\begin{align}
    \mathcal{L}_{\text{pred}} &= \text{MSE}(d_t, \hat{d}_t)\\
    \mathcal{L}_{\text{\componentname}} &= \lambda_z \mathcal{L}_z + \lambda_d \mathcal{L}_d + \lambda_\text{pred} \mathcal{L}_{\text{pred}}
\end{align}
Crucially, the scalar hyperparameters $\lambda_z$ and $\lambda_d$ control the category of information encoded in $z$.
Setting $\lambda_{d}=0$ allows $d$ to collapse and $\text{DiffPred}_{\beta}$ to trivially predict a constant, imposing no additional dynamic feature embeddings on $z_t$ and $z_{t+1}$, which resembles LeWM~\cite{LeWorldModel}.
By setting $\lambda_{z}=0$, we maximize the information on embedded dynamics, creating a model that focuses on motions extracted from the temporal difference image.
Because $\text{DiffPred}_\beta(z_t, z_{t+1})$ predicts the embedding of visual changes $\hat{d}_t$, minimizing $\mathcal{L}_{\text{pred}}$ results in both $z_t$ and $z_{t+1}$ learning transition-relevant information, preventing the collapse of the image embedding $z$ without imposing additional constraints on the shape or distribution of the embedding.
Finally, we join the forward prediction objective of JEPA with our \componentname to obtain the total loss function of \modelname:
\begin{align}
    \mathcal{L}_{\text{\modelname}} = \mathcal{L}_{\text{\componentname}} + \text{MSE}(z_{t+1}, \hat{z}_{t+1})
\end{align}
We provide a pseudocode implementation of the \componentname computation in \cref{alg:disreg}, using the default weights applied throughout our experiments.

\begin{table*}[h!]
    \centering
    \footnotesize
    \begin{subtable}[t]{\linewidth}
    \centering
    \caption{Offline Residual MLP Probe NMSE$\downarrow$ of Image Embedding $z$}
    \begin{tabular}{@{}lc@{\hskip 4pt}c@{\hskip 4pt}c@{\hskip 4pt}cc@{\hskip 4pt}c@{\hskip 4pt}c@{\hskip 4pt}c@{\hskip 4pt}c@{\hskip 4pt}cc@{\hskip 4pt}c@{\hskip 4pt}c@{\hskip 4pt}c@{\hskip 4pt}c@{}}
    \toprule
    & \multicolumn{4}{c}{Pong} & \multicolumn{6}{c}{Dino} & \multicolumn{5}{c}{Golf}\\
    Method & avg. & ball & bats & score & avg. & cactus & clouds & dino & hearts & sky & avg. & bars & cross & ball & wood\\
    \midrule
    ForwardOnly & $1.058$ & $\textcolor{red!70!black}{1.260}$ & $\textcolor{red!70!black}{1.001}$ & $\textcolor{red!70!black}{0.930}$ & $0.949$ & $\textcolor{red!70!black}{1.003}$ & $\textcolor{red!70!black}{1.002}$ & $\textcolor{red!70!black}{1.002}$ & $\textcolor{red!70!black}{1.031}$ & $\textcolor{red!70!black}{0.511}$ & $1.980$ & $\textcolor{red!70!black}{1.233}$ & $\textcolor{red!70!black}{1.558}$ & $\textcolor{red!70!black}{1.077}$ & $\textcolor{red!70!black}{4.324}$\\
    Random & $0.116$ & $\textcolor{red!70!black}{0.143}$ & $\textcolor{green!60!black}{0.093}$ & $\textcolor{red!70!black}{0.115}$ & $0.361$ & $\textcolor{red!70!black}{0.833}$ & $\textcolor{red!70!black}{0.535}$ & $\textcolor{red!70!black}{0.116}$ & $\textcolor{green!60!black}{0.031}$ & $\textcolor{green!60!black}{\mathbf{0.004}}$ & $0.776$ & $\textcolor{red!70!black}{0.857}$ & $\textcolor{red!70!black}{0.296}$ & $\textcolor{red!70!black}{0.989}$ & $\textcolor{red!70!black}{0.653}$\\
    \midrule
    SMWM~\cite{SMWM} & $0.657$ & $\textcolor{red!70!black}{1.001}$ & $\textcolor{green!60!black}{\underline{0.002}}$ & $\textcolor{red!70!black}{1.002}$ & $0.649$ & $\textcolor{red!70!black}{0.434}$ & $\textcolor{red!70!black}{1.075}$ & $\textcolor{green!60!black}{0.028}$ & $\textcolor{red!70!black}{0.787}$ & $\textcolor{green!60!black}{0.071}$ & $0.727$ & $\textcolor{red!70!black}{1.000}$ & $\textcolor{red!70!black}{0.986}$ & $\textcolor{green!60!black}{\mathbf{0.002}}$ & $\textcolor{red!70!black}{1.003}$\\
    LeWM~\cite{LeWorldModel} & $0.658$ & $\textcolor{red!70!black}{0.999}$ & $\textcolor{red!70!black}{1.002}$ & $\textcolor{green!60!black}{\mathbf{0.003}}$ & $0.394$ & $\textcolor{red!70!black}{0.769}$ & $\textcolor{red!70!black}{0.248}$ & $\textcolor{red!70!black}{0.983}$ & $\textcolor{green!60!black}{0.012}$ & $\textcolor{green!60!black}{0.005}$ & $0.763$ & $\textcolor{red!70!black}{1.002}$ & $\textcolor{red!70!black}{1.001}$ & $\textcolor{red!70!black}{0.942}$ & $\textcolor{green!60!black}{0.084}$\\
    LeWM-Time & $1.077$ & $\textcolor{red!70!black}{1.100}$ & $\textcolor{red!70!black}{1.071}$ & $\textcolor{red!70!black}{1.063}$ & $0.913$ & $\textcolor{red!70!black}{1.003}$ & $\textcolor{red!70!black}{0.992}$ & $\textcolor{red!70!black}{1.002}$ & $\textcolor{red!70!black}{1.029}$ & $\textcolor{red!70!black}{0.248}$ & $1.053$ & $\textcolor{red!70!black}{1.008}$ & $\textcolor{red!70!black}{1.014}$ & $\textcolor{red!70!black}{1.004}$ & $\textcolor{red!70!black}{1.195}$\\
    LeWM-Detached & $0.171$ & $\textcolor{red!70!black}{0.174}$ & $\textcolor{red!70!black}{0.331}$ & $\textcolor{green!60!black}{0.007}$ & $0.645$ & $\textcolor{red!70!black}{1.002}$ & $\textcolor{red!70!black}{0.727}$ & $\textcolor{red!70!black}{1.002}$ & $\textcolor{green!60!black}{0.017}$ & $\textcolor{green!60!black}{0.006}$ & $0.811$ & $\textcolor{red!70!black}{0.924}$ & $\textcolor{red!70!black}{0.683}$ & $\textcolor{red!70!black}{0.930}$ & $\textcolor{red!70!black}{0.572}$\\
    LeWM-Flat & $0.284$ & $\textcolor{red!70!black}{0.890}$ & $\textcolor{green!60!black}{0.014}$ & $\textcolor{green!60!black}{0.005}$ & $0.331$ & $\textcolor{red!70!black}{0.522}$ & $\textcolor{red!70!black}{0.593}$ & $\textcolor{green!60!black}{\mathbf{0.002}}$ & $\textcolor{green!60!black}{\mathbf{0.006}}$ & $\textcolor{green!60!black}{0.007}$ & $0.694$ & $\textcolor{red!70!black}{1.002}$ & $\textcolor{red!70!black}{0.379}$ & $\textcolor{red!70!black}{0.993}$ & $\textcolor{green!60!black}{0.052}$\\
    \midrule
    SMWM* & $0.651$ & $\textcolor{red!70!black}{0.983}$ & $\textcolor{green!60!black}{\mathbf{0.002}}$ & $\textcolor{red!70!black}{1.003}$ & $0.606$ & $\textcolor{red!70!black}{0.398}$ & $\textcolor{red!70!black}{1.005}$ & $\textcolor{green!60!black}{0.003}$ & $\textcolor{red!70!black}{0.775}$ & $\textcolor{green!60!black}{0.069}$ & $0.727$ & $\textcolor{red!70!black}{1.000}$ & $\textcolor{red!70!black}{0.986}$ & $\textcolor{green!60!black}{\underline{0.002}}$ & $\textcolor{red!70!black}{1.003}$\\
    LeWM* & $0.061$ & $\textcolor{red!70!black}{0.163}$ & $\textcolor{green!60!black}{0.018}$ & $\textcolor{green!60!black}{0.012}$ & $\underline{0.282}$ & $\textcolor{red!70!black}{0.990}$ & $\textcolor{green!60!black}{\underline{0.059}}$ & $\textcolor{red!70!black}{0.751}$ & $\textcolor{green!60!black}{0.011}$ & $\textcolor{green!60!black}{0.012}$ & $0.760$ & $\textcolor{red!70!black}{1.001}$ & $\textcolor{red!70!black}{1.002}$ & $\textcolor{red!70!black}{0.909}$ & $\textcolor{red!70!black}{0.110}$\\
    LeWM-Time* & $0.687$ & $\textcolor{green!60!black}{0.022}$ & $\textcolor{red!70!black}{0.972}$ & $\textcolor{red!70!black}{1.002}$ & $0.697$ & $\textcolor{green!60!black}{\mathbf{0.005}}$ & $\textcolor{red!70!black}{1.004}$ & $\textcolor{green!60!black}{0.003}$ & $\textcolor{red!70!black}{0.971}$ & $\textcolor{red!70!black}{1.004}$ & $0.985$ & $\textcolor{red!70!black}{0.972}$ & $\textcolor{red!70!black}{0.992}$ & $\textcolor{red!70!black}{0.984}$ & $\textcolor{red!70!black}{1.002}$\\
    LeWM-Detached* & $0.048$ & $\textcolor{red!70!black}{0.103}$ & $\textcolor{green!60!black}{0.038}$ & $\textcolor{green!60!black}{0.008}$ & $0.332$ & $\textcolor{red!70!black}{1.005}$ & $\textcolor{green!60!black}{0.073}$ & $\textcolor{red!70!black}{0.959}$ & $\textcolor{green!60!black}{0.011}$ & $\textcolor{green!60!black}{0.017}$ & $0.745$ & $\textcolor{red!70!black}{1.002}$ & $\textcolor{red!70!black}{1.002}$ & $\textcolor{red!70!black}{0.908}$ & $\textcolor{green!60!black}{\underline{0.048}}$\\
    LeWM-Flat* & $\underline{0.008}$ & $\textcolor{green!60!black}{\underline{0.012}}$ & $\textcolor{green!60!black}{0.004}$ & $\textcolor{green!60!black}{0.008}$ & $0.301$ & $\textcolor{red!70!black}{0.650}$ & $\textcolor{red!70!black}{0.498}$ & $\textcolor{green!60!black}{\underline{0.002}}$ & $\textcolor{green!60!black}{0.007}$ & $\textcolor{green!60!black}{0.016}$ & $\underline{0.299}$ & $\textcolor{red!70!black}{\underline{0.317}}$ & $\textcolor{green!60!black}{\mathbf{0.059}}$ & $\textcolor{red!70!black}{0.312}$ & $\textcolor{red!70!black}{0.377}$\\
    \midrule
    \textbf{\modelname (Ours)} & $\mathbf{0.004}$ & $\textcolor{green!60!black}{\mathbf{0.005}}$ & $\textcolor{green!60!black}{0.003}$ & $\textcolor{green!60!black}{\underline{0.004}}$ & $\mathbf{0.021}$ & $\textcolor{green!60!black}{\underline{0.011}}$ & $\textcolor{green!60!black}{\mathbf{0.037}}$ & $\textcolor{green!60!black}{0.006}$ & $\textcolor{green!60!black}{\underline{0.006}}$ & $\textcolor{green!60!black}{\underline{0.004}}$ & $\mathbf{0.061}$ & $\textcolor{green!60!black}{\mathbf{0.095}}$ & $\textcolor{green!60!black}{\underline{0.083}}$ & $\textcolor{green!60!black}{0.033}$ & $\textcolor{green!60!black}{\mathbf{0.033}}$\\
    \bottomrule
    \end{tabular}
    \end{subtable}
    \vspace{0.5em}

    \begin{subtable}[t]{\linewidth}
    \centering
    \caption{Offline Residual MLP Probe NMSE$\downarrow$ of Predicted Rollout $\hat{z}$}
    \begin{tabular}{@{}lc@{\hskip 4pt}c@{\hskip 4pt}c@{\hskip 4pt}cc@{\hskip 4pt}c@{\hskip 4pt}c@{\hskip 4pt}c@{\hskip 4pt}c@{\hskip 4pt}cc@{\hskip 4pt}c@{\hskip 4pt}c@{\hskip 4pt}c@{\hskip 4pt}c@{}}
    \toprule
    & \multicolumn{4}{c}{Pong} & \multicolumn{6}{c}{Dino} & \multicolumn{5}{c}{Golf}\\
    Method & avg. & ball & bats & score & avg. & cactus & clouds & dino & hearts & sky & avg. & bars & cross & ball & wood\\
    \midrule
    ForwardOnly & $1.010$ & $1.041$ & $1.006$ & $0.992$ & $1.081$ & $1.019$ & $1.006$ & $1.004$ & $1.017$ & $1.634$ & $1.003$ & $1.002$ & $1.001$ & $1.005$ & $1.005$\\
    Random & $0.554$ & $0.721$ & $0.478$ & $0.507$ & $0.617$ & $0.974$ & $0.764$ & $0.631$ & $0.279$ & $0.026$ & $0.934$ & $0.995$ & $0.913$ & $0.998$ & $0.781$\\
    \midrule
    SMWM~\cite{SMWM} & $0.630$ & $1.001$ & $\underline{0.002}$ & $1.002$ & $0.686$ & $0.839$ & $1.002$ & $0.092$ & $0.933$ & $0.076$ & $0.729$ & $1.001$ & $1.001$ & $\mathbf{0.006}$ & $1.002$\\
    LeWM~\cite{LeWorldModel} & $0.644$ & $1.002$ & $1.002$ & $\mathbf{0.002}$ & $0.407$ & $0.856$ & $0.251$ & $0.995$ & $0.017$ & $\mathbf{0.005}$ & $0.767$ & $1.002$ & $1.001$ & $0.947$ & $0.094$\\
    LeWM-Time & $1.001$ & $1.003$ & $0.998$ & $1.002$ & $1.106$ & $1.011$ & $1.009$ & $1.003$ & $1.026$ & $1.845$ & $1.003$ & $1.006$ & $1.006$ & $1.000$ & $1.002$\\
    LeWM-Detached & $0.239$ & $0.397$ & $0.347$ & $0.007$ & $0.646$ & $0.988$ & $0.732$ & $1.002$ & $0.028$ & $0.006$ & $0.855$ & $0.967$ & $0.821$ & $0.930$ & $0.618$\\
    LeWM-Flat & $0.278$ & $0.968$ & $0.039$ & $0.005$ & $0.396$ & $0.877$ & $0.633$ & $0.055$ & $\underline{0.009}$ & $0.007$ & $0.722$ & $1.002$ & $0.590$ & $0.994$ & $0.061$\\
    \midrule
    SMWM* & $0.630$ & $1.001$ & $\mathbf{0.001}$ & $1.002$ & $0.671$ & $0.801$ & $0.999$ & $0.075$ & $0.878$ & $0.079$ & $0.729$ & $1.001$ & $1.001$ & $\underline{0.006}$ & $1.002$\\
    LeWM* & $0.138$ & $0.471$ & $0.020$ & $0.009$ & $\underline{0.334}$ & $0.991$ & $\underline{0.060}$ & $1.003$ & $0.018$ & $0.012$ & $0.763$ & $1.002$ & $1.002$ & $0.913$ & $0.117$\\
    LeWM-Time* & $0.868$ & $0.507$ & $1.002$ & $1.002$ & $0.710$ & $\mathbf{0.120}$ & $1.004$ & $\mathbf{0.020}$ & $0.959$ & $1.003$ & $0.997$ & $0.997$ & $0.995$ & $0.991$ & $1.002$\\
    LeWM-Detached* & $0.129$ & $0.395$ & $0.053$ & $0.008$ & $0.343$ & $0.988$ & $0.078$ & $1.002$ & $0.020$ & $0.014$ & $0.746$ & $1.002$ & $1.001$ & $0.909$ & $\underline{0.051}$\\
    LeWM-Flat* & $\underline{0.054}$ & $\underline{0.183}$ & $0.004$ & $0.007$ & $0.353$ & $0.923$ & $0.524$ & $0.066$ & $0.013$ & $0.018$ & $\underline{0.448}$ & $\underline{0.540}$ & $\mathbf{0.320}$ & $0.402$ & $0.425$\\
    \midrule
    \textbf{\modelname (Ours)} & $\mathbf{0.036}$ & $\mathbf{0.122}$ & $0.004$ & $\underline{0.005}$ & $\mathbf{0.055}$ & $\underline{0.243}$ & $\mathbf{0.040}$ & $\underline{0.051}$ & $\mathbf{0.007}$ & $\underline{0.005}$ & $\mathbf{0.164}$ & $\mathbf{0.262}$ & $\underline{0.348}$ & $0.065$ & $\mathbf{0.037}$\\
    \bottomrule
    \end{tabular}
    \end{subtable}

    \caption{\textbf{Probing Results on Game Environments}. We use high-capacity MLP probes with residual connections to probe the latent embedding $z$ for predicting environment features and report the NMSE averaged over three runs. (a) Probes the image embedding directly from the encoder; (b) probes on rolled-out states of a rollout length of 5. Runs marked with * are from a hyperparameter search. For (a), based on the results of the random weights model on Pong, NMSE values $\leq 0.1$ are marked in \textcolor{green!60!black}{green} as not collapsed; otherwise, \textcolor{red!70!black}{red} for collapsed. Best results in each column are highlighted in \textbf{bold} and second best \underline{underlined}.}
    \label{tab:games_offline_mlp_probes}
\end{table*}

\begin{figure*}
  \centering{\includegraphics[width=\textwidth]{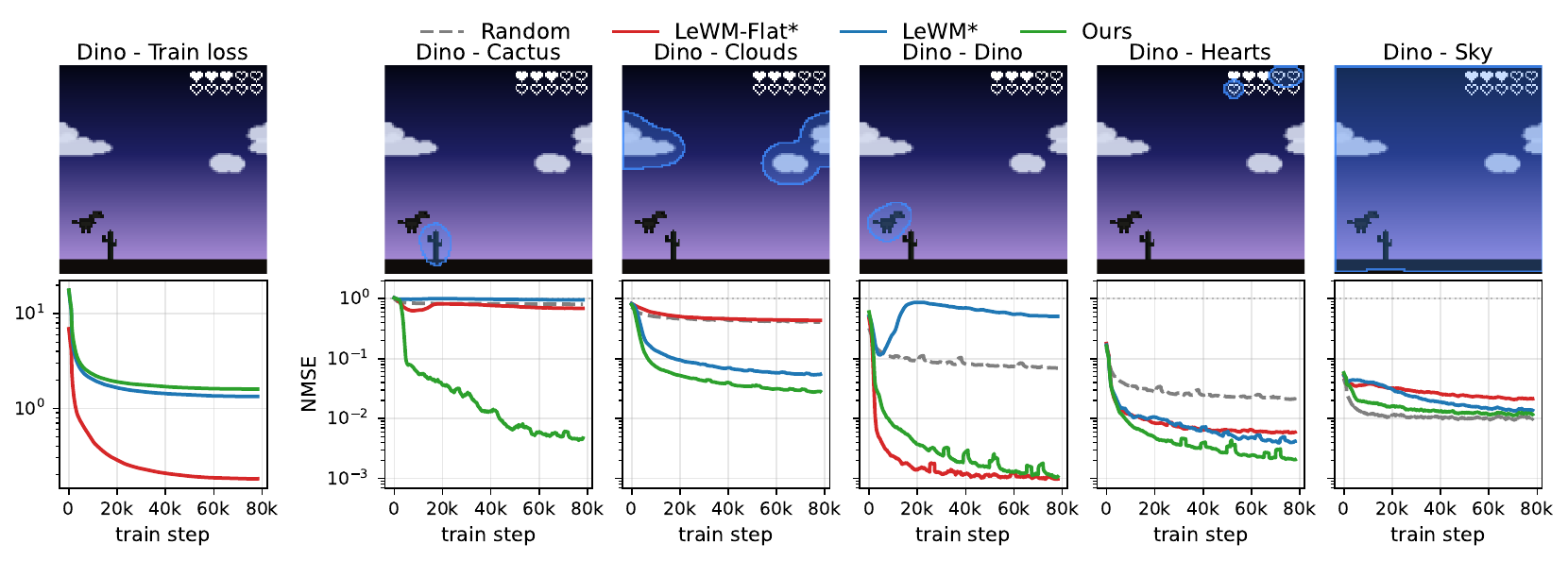}}
  \caption{\textbf{State Probing during Training}. We train a detached state decoder alongside training the JEPA model. The first column of each environment is the training loss (log scale). The remaining curves are per-feature probe NMSE. Image areas affected by the corresponding feature are highlighted in blue. Models marked with * are optimized with a hyperparameter search.}
  \label{fig:games_online_mlp_probes_during_training}
\end{figure*}

\section{Experiments} \label{sec:experiments}
The experiments are organized into two main sections. 
\cref{sec:suppression_and_collapse_analysis} investigates collapse scenarios on game environments with slow and static features by probing the latent embedding $z$ and further analyzing its path straightness. 
\cref{sec:downstream_tasks} demonstrates our method's performance on downstream tasks with static distractors. 
Throughout all experiments, we use the same set of hyperparameters for \modelname's \componentname: $\lambda_{\mathrm{z}}=0.25$, $\lambda_{\mathrm{pred}}=0.5$, and $\lambda_{\mathrm{d}}=2$.

\paragraph{Baselines.}
We compare our method with previous end-to-end JEPA-style latent world models, namely \textbf{LeWM}~\cite{LeWorldModel} and \textbf{SMWM}~\cite{SMWM} with default SIGReg weight of $0.09$ for LeWM and inverse prediction weight of $1.0$ for SMWM.
We additionally provide altered versions of LeWM with common practice modifications used in the literature to improve robustness against feature collapse. 
\textbf{LeWM-Time} applies SIGReg along the time dimension~\cite{LeNEPA}, \textbf{LeWM-Flat} applies SIGReg along the flattened batch and time dimension, and \textbf{LeWM-Detached} adds a StopGradient operator in the JEPA predictor loss target embedding $\text{MSE}(\text{StopGradient}(z_{t+1}), \hat{z}_{t+1})$~\cite{BYOL, IJEPA, VJEPA}.
Besides these baselines, we also include \textbf{ForwardOnly}, which has only the forward prediction loss and no anti-collapse mechanism, and \textbf{Random}, with fixed, randomly initialized weights, showing what information is present in the latent at initialization that might be suppressed during training.
Finally, we also perform a hyperparameter search for each baseline environment combination to demonstrate that the collapse is not an issue of hyperparameter selection but a fundamental limitation of JEPA anti-collapse mechanisms~\cite{JEPAsFocusOnSlowFeatures}.
We mark such models with a star * throughout this work.

\subsection{Collapse and Latent Embedding Analysis} \label{sec:suppression_and_collapse_analysis}

\paragraph{Environments.}
We designed three fully synthetic, deterministic game environments with clearly separable feature groups.
Samples for each environment are generated on the fly during training and evaluation to prevent overfitting\footnote{Sampling is seeded to ensure that all methods observe identical training and evaluation sequences for a given seed.}.
Initial states and actions are sampled randomly for each episode to minimize correlations between features.
This reduces the risk of shortcut learning~\cite{ShortcutLearning}; for example, in Pong, a probe estimating the bat's position solely from the ball's location would falsely suggest that bat information is encoded when it is not.

We propose three environments of increasing dynamical complexity: \textbf{Pong}, where the agent controls both bats to hit the ball; \textbf{Dino}, a 2D side-scroller where an agent must jump over a cactus; and \textbf{Golf}, in which the player navigates a ball toward a hole while avoiding moving obstacles.
We design the true state dimension of each environment (excluding velocities: Pong: 6, Dino: 8, Golf: 6) to be significantly lower than the latent embedding dimension $D_z$ of $192$ to ensure that the latent dimensionality is not a bottleneck that artificially induces collapse.
Further implementation details for each environment are provided in \cref{sec:appendix_implementation_details}.

\paragraph{Evaluation Setup.}
We follow \cite{JEPAsFocusOnSlowFeatures} and analyze the collapse of features by probing the learned image embedding $z$.
To keep the comparison fair, all methods use the same image encoder, latent forward predictor, training schedule, and optimizer from \cite{LeWorldModel}.
After training from scratch on 10M randomly sampled episodes, we freeze the base model, train probes on 1M training samples, and evaluate those probes on 10k episodes.

For the collapse analysis, we choose higher-capacity MLP probes with residual connections instead of linear probe, as we are not interested in the linear separability or ease of accessibility of features, but rather in measuring the general presence or suppression of information in the latent embedding $z$.
For example, a world model which uses a non-linear, highly correlated representation in $z$ might still be functional, yet linear probes might fail to decode that information.

We report Normalized Mean Squared Error (NMSE) for each probe.
An NMSE of $1.0$ corresponds to predicting the evaluation target mean, indicating a complete failure to extract useful information from $z$.
We train and evaluate each method-environment pair for three seeds and report the average results.
We provide more details about the training and evaluation process in \cref{sec:appendix_implementation_details}.

\begin{figure*}
  \centering{\includegraphics[width=\textwidth]{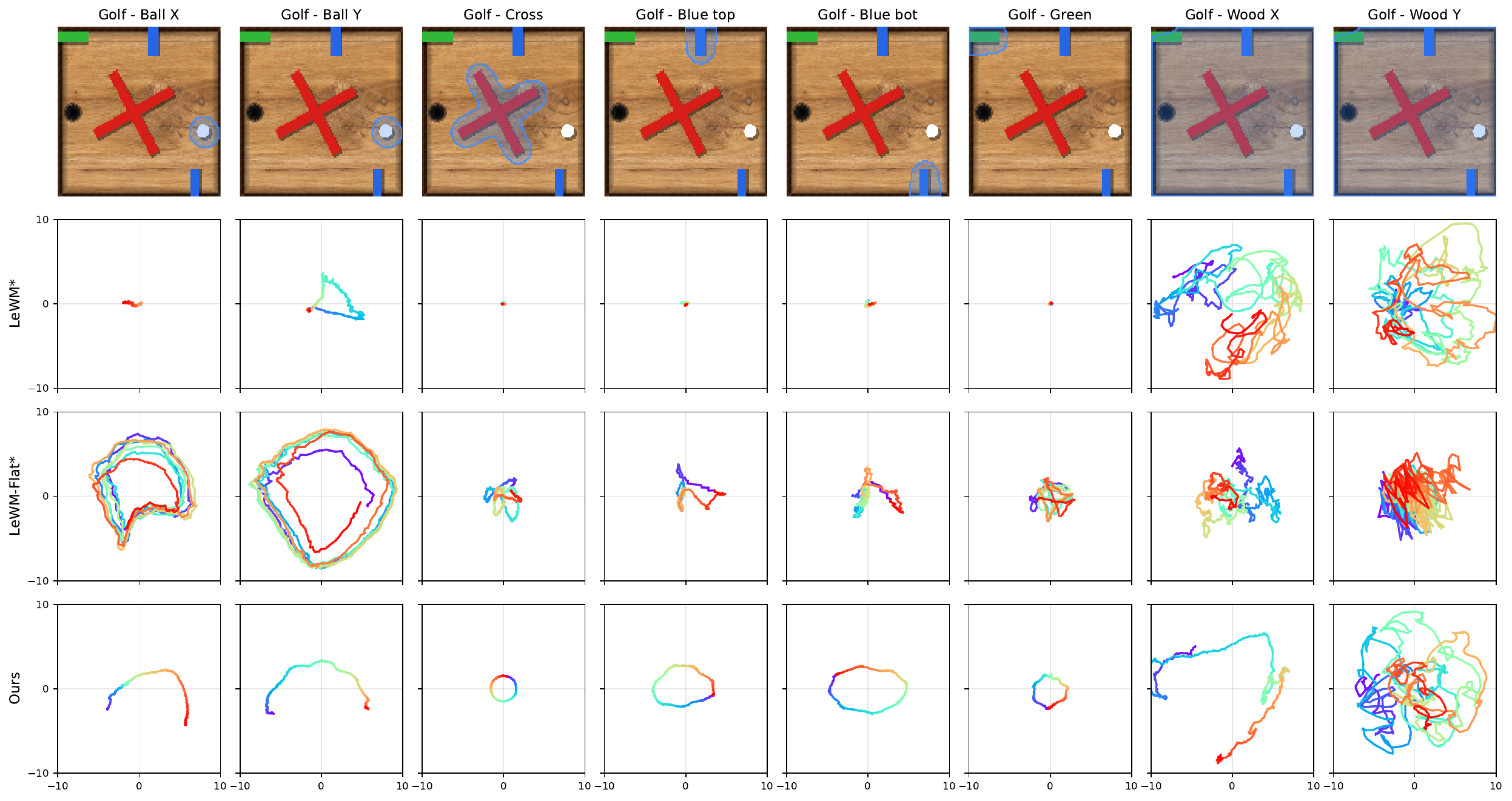}}
  \caption{\textbf{Latent Trajectories}. We visualize how individual state features map into the latent space. For each subplot, we vary a single feature across a 1D range while holding the remaining state variables fixed, render the resulting sequence into images, and encode them into latent representations $z$. We then project these latent trajectories onto their first two principal components without variance rescaling. We further color the state range in rainbow colors from \textcolor[HTML]{7F00FF}{$0$}\textcolor[HTML]{6300E6}{:}\textcolor[HTML]{4500CC}{~}\textcolor[HTML]{2200B3}{v}\textcolor[HTML]{0015A8}{i}\textcolor[HTML]{0039B8}{o}\textcolor[HTML]{005CC8}{l}\textcolor[HTML]{007FD8}{e}\textcolor[HTML]{0099C0}{t}\textcolor[HTML]{009A8E}{~}\textcolor[HTML]{008C50}{t}\textcolor[HTML]{007E1A}{o}\textcolor[HTML]{3D7A00}{~}\textcolor[HTML]{737A00}{$1$}\textcolor[HTML]{A87400}{:}\textcolor[HTML]{D46200}{~}\textcolor[HTML]{E84800}{r}\textcolor[HTML]{F42400}{e}\textcolor[HTML]{FF0000}{d}.}
  \label{fig:games_latent_trajectory_main_section}
\end{figure*}

\paragraph{Probing Results.}
\cref{tab:games_offline_mlp_probes} shows the MLP probing results on Pong, Dino, and Golf for both the encoder embeddings $z$ as well as predicted rollouts $\hat{z}$.
The goal is to obtain the lowest average loss across all categories for each environment, indicating stable and balanced learning of all features\footnote{The average column presents a weighted mean, where each feature group's weight corresponds to its dimensionality.}.
Looking at the results of probing the image embedding $z$, as expected, the control method ForwardOnly collapsed on all three tasks, resulting in an NMSE greater than or about $1.0$ for all features except the sky color of Dino, with $0.511$, which is arguably the easiest feature of these environments.
Switching to the Random model, we observe that even though the weights are random, the probes are still able to extract a significant amount of information, with an NMSE average of $ 0.116$ for Pong and even $0.004$ for the sky in the Dino environment.
This shows that even though the feature information is accessible at the start of training, models actively suppress it during training.

Using default training parameters for LeWM results in clear collapses across all three environments, focusing only on slow or static features such as the score in Pong.
While the hyperparameter-optimized version LeWM* improves the average probing loss on a simple task such as Pong from $0.658$ to $0.061$, in the presence of more dynamic features, as in Dino and Golf, such hyperparameter optimization does not significantly improve results (see grid search in the appendix in \cref{fig:appendix_hp_grid_search}), highlighting a core limitation of not regularizing along the time dimension.

Both the LeWM-Flat* and LeWM-Time* variants add distribution constraints on the time dimension.
LeWM-Time is the only model variant besides ours that successfully encodes the cactus in Dino, yet completely fails in the presence of more complex moving features in Golf.
SMWM uses an inverse dynamics module that predicts the action of two consecutive states.
It performs well on states which are directly impacted by actions but collapses on other task-relevant information.

Our method achieves the lowest average feature probing loss across all three environments for both image embedding and rollout prediction while using the \textbf{same set of hyperparameters} across all three environments.
\modelname is also the only method that achieves a balanced probing loss below $0.1$ on all categories, indicating robustness against feature collapse.
Especially for the bar obstacle category of Golf, where \modelname obtains $0.095$ while all other methods suppress this category with losses higher than $0.3$.
We also provide results with linear probes in \cref{tab:appendix_games_linear_offline_probes} of the appendix, in which our model also achieves the lowest average probing loss across all environments.

\paragraph{When does collapse happen?}
In \cref{fig:games_online_mlp_probes_during_training} we show the training loss curves with additional online MLP probes for the Dino environment.
The probes are trained on the $z$ embedding with a stop-gradient operator to prevent information leaking to the encoder.
The static sky and slow-changing hearts are the easiest features, quickly saturating at about 10k steps to $0.01$ NMSE.
We also observe that the random weights model is encoding these features as well.
For LeWM, in which SIGReg is regularizing the distribution along the batch dimension, the slow features of the sky, hearts, and clouds suppress the embedding of the cactus and dinosaur.
This is especially visible for the dinosaur, as the probing loss first decreases to approximately $0.1$ at 5k iterations and then increases back to $1.0$ at 20k iterations, from which it is not able to recover.

LeWM-Flat, which enforces a distribution across both batch and time dimensions, seeks variance along the time dimension of the embedding as well and therefore quickly picks up the dinosaur.
This is probably because its limited movement of jumping is very simple, but provides consistent temporal change.

Crucially, the collapse is not visible in the training loss (see first column), demonstrating that monitoring loss alone is insufficient to diagnose collapse without explicit latent probing or downstream evaluation.

\paragraph{Latent Path Straightness.}
The temporal straightening hypothesis~\cite{Henaff18} states
that useful representations turn curved input dynamics into near-straight
latent paths. It has proven informative in practice, as a signal for
identifying AI-generated video~\cite{interno2025aigenerated}, for judging physical
plausibility~\cite{internò2026geophysgeometryphysicalplausibility}, and as a planning
objective~\cite{wang2026temporal}.
We show empirical results of the latent path straightness in \cref{tab:path_straightness}.
SMWM achieves the straightest latent trajectories close to the ideal straight-line value of $1.0$ in both Pong and Golf, as it has no distribution regularization on $z$.
Introducing distribution regularization along the time axis, such as in variants LeWM-Time* and
LeWM-Flat* causes the opposite effect, significantly increasing the latent embedding curvature to above $20$ for both Pong and Golf.
As our method is architecturally closer to SMWM's IDM module, we observe straighter paths than all distribution regularization methods on Pong and Golf.

This can also be observed in \cref{fig:games_latent_trajectory_main_section}, where we show example latent trajectories of each feature of the golf environment.
Our method produces nearly straight latent embeddings for ball position and single-loop embeddings for the circular cross and bar obstacles\footnote{The bar obstacles reappear on the opposite side if exiting the image and therefore result in a looped trajectory.}.
LeWM collapses all obstacles into almost a single point while showing high-curvature embeddings in the distracting wood background.
While the representations of the LeWM-Flat variant do not collapse into a single point, the embedding has more curvature and entanglement, making it difficult to separate and predict individual features.
We provide additional latent trajectory visualization of other environments and methods in the appendix.

\begin{table}[h!]
    \centering
    \small
    \begin{tabular}{@{}lccc@{}}
    \toprule
    Method & Pong & Dino & Golf\\
    \midrule
    ForwardOnly & $45.514$ & $\textbf{4.045}$ & $26.412$\\
    Random & $25.278$ & $13.660$ & $26.842$\\
    \midrule
    SMWM~\cite{SMWM} & $\textbf{1.207}$ & $7.557$ & $\textbf{1.078}$\\
    LeWM~\cite{LeWorldModel} & $5.170$ & $\underline{4.826}$ & $8.471$\\
    LeWM-Time & $29.911$ & $4.955$ & $27.190$\\
    LeWM-Detached & $3.067$ & $18.372$ & $15.455$\\
    LeWM-Flat & $3.031$ & $12.986$ & $5.298$\\
    \midrule
    SMWM* & $\underline{1.427}$ & $8.631$ & $\underline{1.078}$\\
    LeWM* & $8.170$ & $7.755$ & $8.191$\\
    LeWM-Time* & $23.376$ & $9.979$ & $38.839$\\
    LeWM-Detached* & $4.111$ & $8.806$ & $7.935$\\
    LeWM-Flat* & $22.159$ & $15.610$ & $26.663$\\
    \midrule
    \textbf{\modelname (Ours)} & $2.756$ & $5.789$ & $3.305$\\
    \bottomrule
    \end{tabular}

    \caption{\textbf{Latent path straightness}. We calculate the mean encoder arc length over the chord length for 100 random linear state-space interpolations. Lower is straighter, with 1.0 being a straight line.}
    \label{tab:path_straightness}
\end{table}

\subsection{Downstream Tasks}\label{sec:downstream_tasks}
As in LeWorldModel~\cite{LeWorldModel}, probing is intended primarily as an analysis of representation content rather than to substitute for closed-loop control evaluation.
We therefore take the experimental setup of LeWM~\cite{LeWorldModel} and evaluate the model's success rate in the presence of slow features.

\begin{figure}[t]
    \centering
    \begin{subfigure}[t]{0.24\linewidth}
        \centering
        \includegraphics[width=\linewidth]{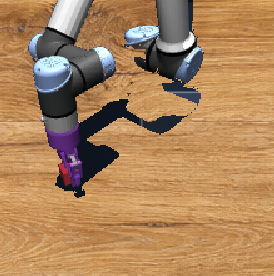}
        \caption{Cube}
    \end{subfigure}
    \hfill
    \begin{subfigure}[t]{0.24\linewidth}
        \centering
        \includegraphics[width=\linewidth]{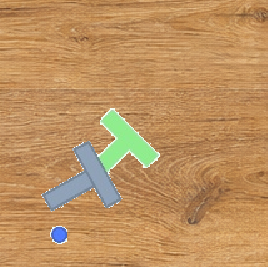}
        \caption{PushT}
    \end{subfigure}
    \hfill
    \begin{subfigure}[t]{0.24\linewidth}
        \centering
        \includegraphics[width=\linewidth]{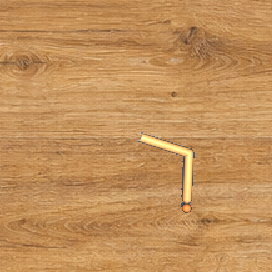}
        \caption{Reacher}
    \end{subfigure}
    \hfill
    \begin{subfigure}[t]{0.24\linewidth}
        \centering
        \includegraphics[width=\linewidth]{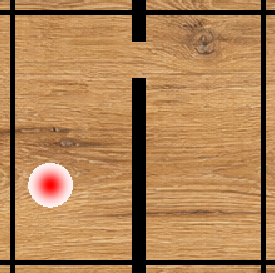}
        \caption{TwoRoom}
    \end{subfigure}

    \caption{Example observations from the four LeWorldModel environments with a wood-texture background used as a static visual distractor. A different crop of the wood texture is sampled for each rollout while remaining fixed within the rollout.}
    \label{fig:lewm_wood}
\end{figure}

\paragraph{Environments.}
All four environments originally contain no static or slow visual features.
To introduce a static-background distractor, we replace the original background with a wood-texture image, mimicking games played on a wooden table.
For each rollout, a different crop of the texture image is sampled and kept fixed throughout the rollout. For PushT and TwoRoom, the original white background pixels are directly replaced with the wood texture.
For Reacher and OG-Cube, we first estimate a static-background mask using the pixel-wise median over the training dataset and replace the corresponding regions with the texture. Example observations from all four environments are shown in Fig.~\ref{fig:lewm_wood}.
All models are trained for 10 epochs with a batch size of 64 using AdamW, with a learning rate of $5\times10^{-5}$ and weight decay of $10^{-3}$.

\paragraph{Evaluation and Results.}
For evaluation, we use CEM-based planning following the protocol of the original LeWorldModel paper~\cite{LeWorldModel}. 
For IDM, we use our own implementation of inverse-dynamics supervision, predicting the raw action from $\Delta z_t=z_{t+1}-z_t$ with a three-layer MLP. This follows the $\Delta z$ input formulation of DeltaJEPA~\cite{DeltaJEPA}, while SMWM~\cite{SMWM} similarly employs inverse-dynamics supervision with a different latent input parameterization.
As an additional baseline, we evaluate LeWM-Flat, which applies SigReg after flattening the batch and temporal dimensions.
Initial states are randomly sampled from the test split, and we evaluate planning horizons of 25 and 50 steps. 
For the 25-step setting, we follow the CEM planning configuration of \cite{LeWorldModel}, using 300 candidate sequences, 30 elites, and the same environment-specific optimization budget. With a frame skip of 5, CEM plans five model steps ahead, corresponding to 25 environment steps.
For the 50-step setting, we first plan and execute a 25-step trajectory, then replan from the resulting state for the remaining 25 steps using the same CEM configuration.
The results are summarized in Table~\ref{tab:lewm_results}.
Overall, the original LeWM baseline degrades substantially in the presence of the static-background distractor, particularly on PushT and Reacher, while LeWM-Flat provides a stronger baseline but still remains clearly below IDM and our method in mean performance.
IDM improves robustness across the four environments, while our standalone method achieves stronger overall mean performance, with particularly large gains on Cube and Reacher under the longer 50-step planning horizon.
Combining our objective with IDM yields the best overall results, further indicating that the two training signals are complementary.
On TwoRoom, both approaches achieve near-saturated performance, whereas the original LeWM baseline remains considerably lower.
We further probe task-relevant states from the frozen representations on the four LeWorldModel benchmarks with static wood-texture distractors.
Our method substantially improves state recoverability over LeWorldModel, consistent with its stronger downstream planning performance.
Detailed linear/MLP probing and reconstruction results are provided in Sec.~\ref{sec:appendix_lewm_probing}.

\begin{table}[t]
\centering
\small
\setlength{\tabcolsep}{3.2pt}
\begin{tabular}{@{}lccccc@{}}
\toprule
Method & Cube & PushT & Reach. &
2Room & Mean \\
\midrule

\multicolumn{6}{c}{25-step planning} \\
\midrule
\multicolumn{6}{l}{\emph{Baseline methods}} \\
LeWM~\cite{LeWorldModel}
& 43.6 & 4.0 & 5.2 & 28.8 & 20.4 \\
LeWM-Flat & 55.2 & 14.4 & 49.2 &  48.0 & 41.7\\
IDM~\cite{DeltaJEPA, SMWM}
& 76.0 & \textbf{83.6} & 49.2 & \textbf{100.0} & 77.2 \\

\addlinespace[2pt]
\multicolumn{6}{l}{\emph{Our method}} \\
Ours ($\lambda_z=0$)
& 60.4 & 76.8 & 24.8 & 99.6 & 65.4 \\
Ours
& \underline{78.0} & 78.4 & \underline{71.2} & 99.6 & \underline{81.8} \\
Ours + IDM
& \textbf{80.8} & \underline{82.4} & \textbf{82.4} & \textbf{100.0} & \textbf{86.4} \\

\midrule
\multicolumn{6}{c}{50-step planning} \\
\midrule
\multicolumn{6}{l}{\emph{Baseline methods}} \\
LeWM~\cite{LeWorldModel}
& 24.4 & 2.4 & 2.0 & 18.4 & 11.8 \\
LeWM-Flat & 42.4 &  2.0 &  60.4 &  22.8 & 31.9\\
IDM~\cite{DeltaJEPA, SMWM}
& 53.2 & 26.0 & 62.0 & \underline{96.0} & 59.3 \\

\addlinespace[2pt]
\multicolumn{6}{l}{\emph{Our method}} \\
Ours ($\lambda_z=0$)
& 48.0 & 21.2 & 33.6 & 94.0 & 49.2 \\
Ours 
& \textbf{63.2} & \underline{28.0} & \underline{98.0} & 92.0 & \underline{70.3} \\
Ours + IDM
& \underline{61.2} & \textbf{29.6} & \textbf{98.4} & \textbf{97.2} & \textbf{71.6} \\
\bottomrule
\end{tabular}
\caption{Planning success rate (\%) under static-background distractors.
Our method gives the best standalone mean performance, while combining it
with IDM yields the best overall results.}
\label{tab:lewm_results}
\end{table}

\paragraph{Ablation Studies.}
For the four LeWM benchmark tasks, our method builds on LeWM and introduces an additional prediction objective using temporal difference images. 
As shown in Table~\ref{tab:lewm_results}, removing the regularization on the original latent representation ($\lambda_z=0$) degrades performance, indicating that preserving global state information remains important alongside change-aware supervision. 
We further apply the same difference-image objective to the IDM-based LeWM variant and observe consistent improvements in 25-step planning, showing that the proposed objective is complementary to inverse-dynamics supervision and can serve as a useful add-on to existing world-model training.
\section{Limitations}\label{sec:limitations}
\componentname introduces two additional modules, 
$\text{DiffEnc}_\alpha$ and $\text{DiffPred}_\beta$, both increasing the parameter count and compute cost relative to a plain JEPA baseline. 
However, the training loss curves in \cref{fig:games_online_mlp_probes_during_training} show that collapse is not resolved simply by training longer or with more capacity.
The baselines plateau in NMSE despite continued training, indicating that feature suppression is not an issue of compute budget.
Furthermore, the modules introduced by \componentname are only relevant during training and are discarded afterward, adding no computational overhead during inference.

\section{Conclusion}
\label{sec:conclusion}
We introduced \componentname, a regularizer that encourages JEPA world models to encode dynamic features by predicting an embedding of the temporal difference image, without action labels or pixel reconstruction.
Integrated in \modelname, our model captures both static and dynamic features more completely than prior methods, with straighter, less entangled latent trajectories, using one fixed hyperparameter set throughout.
This translates to improved downstream planning under static-background distractors.
Overall, encouraging dynamics through prediction rather than direct latent constraints offers a simple, effective route to collapse-resistant JEPA world models.
{
    \small
    \bibliographystyle{ieeenat_fullname}
    \bibliography{main}
}

\clearpage
\setcounter{page}{1}
\maketitlesupplementary

\begin{figure*}
  \begin{subfigure}[t]{\linewidth}
    \caption{Pong}
    \centering{\includegraphics[width=0.9\textwidth]{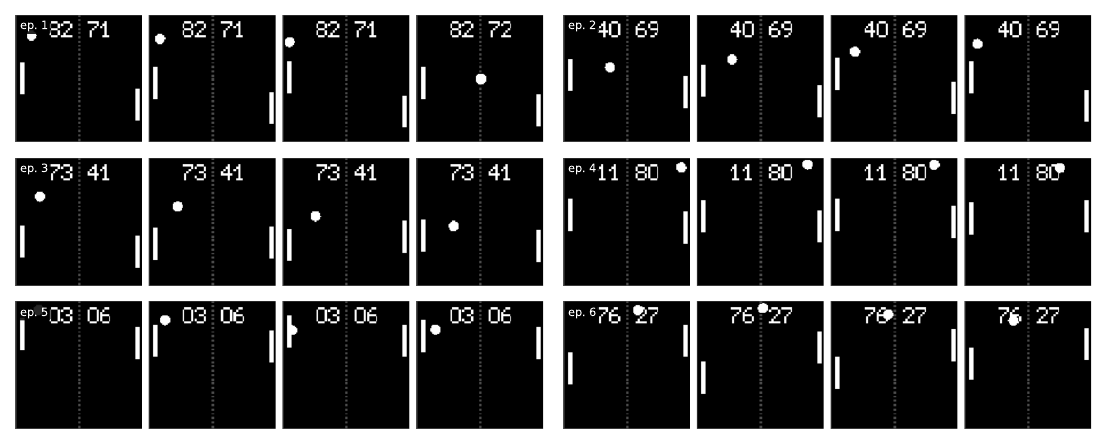}}
  \end{subfigure}
  \begin{subfigure}[t]{\linewidth}
    \caption{Dino}
    \centering{\includegraphics[width=0.9\textwidth]{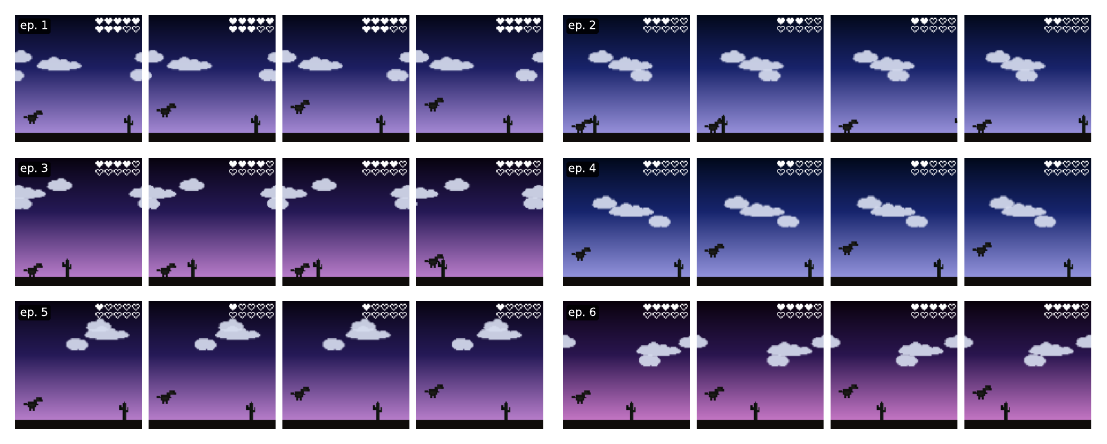}}
  \end{subfigure}
  \begin{subfigure}[t]{\linewidth}
    \caption{Golf}
    \centering{\includegraphics[width=0.9\textwidth]{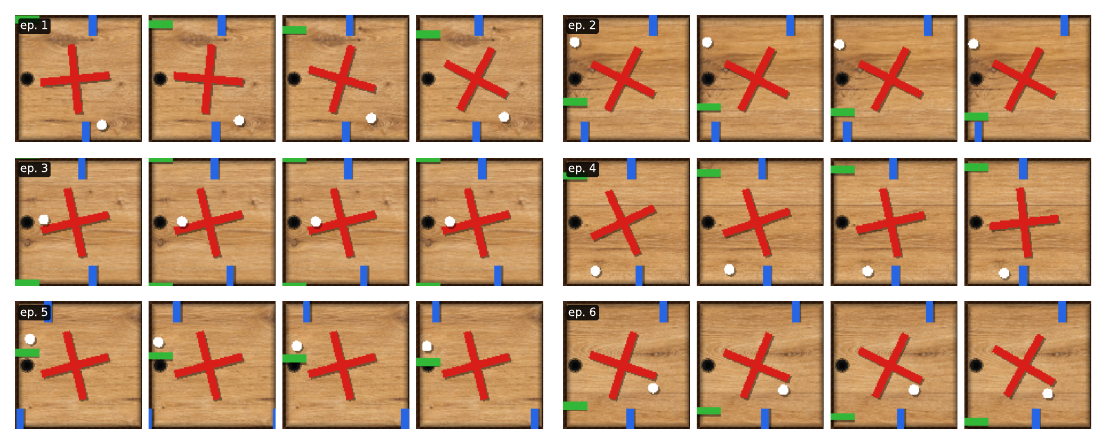}}
  \end{subfigure}
  \caption{\textbf{Randomly sampled episodes of game environments}. We show six randomly sampled 4-frame episodes of each environment. Each episode is the same length as used during training. Episode one of Pong shows the right player scoring, and episode two of Dino demonstrates the player losing one heart by running into the cactus.}
  \label{fig:appendix_random_game_episodes}
\end{figure*}

\section{Implementation Details} \label{sec:appendix_implementation_details}

\subsection{Model and Component Architectures}

\paragraph{Encoder and Predictor Architecture.}
We use the same encoder and predictor architectures as \cite{LeWorldModel}.
The encoder is a ViT-Tiny Vision Transformer with a patch size of $14 \times 14$ pixels.
As in \cite{LeWorldModel}, the $D_z=192$ dimensional embedding vector is obtained from the [CLS] token of the final layer, followed by a 1-layer MLP with Batch Normalization~\cite{BatchNorm}.
The predictor is a six-layer causal transformer with 16 attention heads and learned positional embeddings~\cite{LeWorldModel}.

\paragraph{\componentname Components.}
We use the same architecture for the $\text{DiffEnc}_\alpha$ as for $\text{Enc}_\theta$.
The difference predictor $\text{DiffPred}_{\beta}$ is a three-layer MLP with hidden layers of size 512 that uses LayerNorm and GELU after all but the final linear layer.
The input is a concatenated vector of the two consecutive embeddings $z_t$ and $z_{t+1}$.

\paragraph{Residual MLP Probe.}
The residual MLP probe has the highest capacity and is intended to check the general presence of information, not ease of accessibility.
It is an MLP of depth 6 and width 128 with residual connections between every second layer and pre-activation batch normalization.

\paragraph{MLP Probe.}
We train a two-layer MLP on top of the frozen latent representation to measure non-linear accessibility of the states, comprising a linear layer with 512 hidden units, a GELU activation, and a final linear output layer. Angular targets are represented by their sine and cosine values and trained with MSE; angular error is computed after decoding the predicted angle.

\paragraph{Linear Probe.}
We train a single affine linear layer, including a bias term, on top of the frozen latent representation.
No hidden layers, normalization, or activation functions are used. For LeWM benchmark problems, angular targets are represented by their sine and cosine values and trained with MSE; angular error is computed after decoding the predicted angle.

\subsection{Game Environments} \label{sec:appendix_game_environments}
For all games, the states and actions are normalized to the range $[0, 1]$.
Initial states for the first frame and actions are sampled independently from a uniform distribution.
These states are then simulated deterministically with the given actions to generate a total of 4 states and corresponding frames.
Sample episodes of each dataset are shown in \cref{fig:appendix_random_game_episodes}.

\subsubsection{Pong}
\noindent\colorbox{blue!10}{\parbox{\dimexpr\columnwidth-2\fboxsep\relax}{
\textbf{\textcolor{blue}{State Variables}}: left bat position, right bat position, ball x-position, ball y-position, ball x-velocity, ball y-velocity, score left player, score right player.
}}\\[4pt]
\noindent\colorbox{orange!10}{\parbox{\dimexpr\columnwidth-2\fboxsep\relax}{
\textbf{\textcolor{orange}{Actions}}: left bat velocity, right bat velocity.
}}

\noindent The positions of the ball and the players' bats are also uniformly initialized.
If a player hits the ball, the ball's trajectory changes towards the opponent.
The change in angle is dependent on the bat's position, and the horizontal velocity accelerates with each hit up to a maximum velocity.
If one player scores a goal, the score of the respective player increases by one, and the ball is reset to the center, moving towards the player who was scored against.
The task requires the model to estimate the ball's velocity in order to predict the ball's next position.

\subsubsection{Dino}
\noindent\colorbox{blue!10}{\parbox{\dimexpr\columnwidth-2\fboxsep\relax}{
\textbf{\textcolor{blue}{State Variables}}: dinosaur y-position, dinosaur y-velocity, cactus x-position, cactus x-velocity, normalized number of hearts, sky color, clouds height, cloud 0 scroll, cloud 1 scroll, cloud 2 scroll.
}}\\[4pt]
\noindent\colorbox{orange!10}{\parbox{\dimexpr\columnwidth-2\fboxsep\relax}{
\textbf{\textcolor{orange}{Actions}}: jump.
}}
\noindent A side scroller in which the goal is to time the dinosaur's jumps to avoid upcoming cacti.
Clouds each move with an individual fixed velocity for each episode.
If the dinosaur collides with one of the cacti, the number of hearts is reduced by 1 (corresponding to a change of $0.1$ in the normalized state vector).
Once all lives are depleted, the game freezes for all remaining frames.

\subsubsection{Golf}
\noindent\colorbox{blue!10}{\parbox{\dimexpr\columnwidth-2\fboxsep\relax}{
\textbf{\textcolor{blue}{State Variables}}: ball x-position, ball y-position, cross angle, cross angular velocity, blue bar 1 x-scroll, blue bar 1 scroll x-velocity, blue bar 2 x-scroll, blue bar 2 x-scroll velocity, green bar y-scroll, green bar y-scroll velocity.
}}\\[4pt]
\noindent\colorbox{orange!10}{\parbox{\dimexpr\columnwidth-2\fboxsep\relax}{
\textbf{\textcolor{orange}{Actions}}: ball x-velocity, ball y-velocity.
}}
\noindent In this game, the player navigates a ball through a 2D course of moving obstacles.
Each obstacle is initialized at a random position, together with a random velocity that is fixed for each episode.
The bars, when exiting the screen on one side, reappear on the other side of the screen.
Colliding with obstacles stops the game, freezing it for the remaining frames and ignoring all action inputs.
As the red cross is symmetric to one quarter of a full rotation, we normalize the cross angle to a range of $0$ to $90$ degrees.

\subsection{Details for Training on Game Environments}
Training and probe fitting use mixed-precision {\it bfloat16} automatic mixed precision and {\it torch.compile}.

\paragraph{Pre-training on Game Environments.}
Experiments on the four game environments are conducted at a resolution of $112 \times 112$ pixels with no skip frames (i.e., simple next-frame prediction).
We train the models using a history size of three on 10 million four-frame episodes with a batch size of 128, resulting in approximately 78k training steps.
We use the AdamW optimizer with a weight decay of $10^{-3}$.
For the learning rate, we employ a linear-warmup cosine-annealing schedule, matching the LeWM~\cite{LeWorldModel} defaults: the learning rate rises linearly from $0$ to $5 \times 10^{-5}$ over the first $1\%$ of training (781 steps), then decays to zero following a cosine schedule over the remaining steps.
This schedule is stepped after every optimizer update.

\paragraph{Fitting Probes on Game Environment Models.}
After pre-training, we load and freeze the weights of the resulting encoder $\text{Enc}_\theta$ and predictor $\text{Pred}_\phi$.
We generate 1 million eight-frame episodes and, for each episode, embed the first three frames into latent vectors $z_{1:3}$ using the encoder $\text{Enc}_\theta$.
Next, we use the first three embeddings $z_{1:3}$ as context and apply the predictor $\text{Pred}_\phi$ autoregressively with a sliding window of size three and conditioned on the episode’s ground-truth actions to generate a rollout of five predicted embeddings $\hat{z}_{4:8}$.
Either image embeddings $z_{1:3}$ or predicted rollout embeddings $\hat{z}_{4:8}$ are then used as inputs for the Residual MLP or Linear Probes.
This results in a total of four probes: a Linear Probe on image embeddings, a Linear Probe on rollout embeddings, a Residual MLP Probe on image embeddings, and a Residual MLP Probe on rollout embeddings.
The training target for each probe is the state vector detailed in \cref{sec:appendix_game_environments}.
We use the AdamW optimizer with a weight decay of $10^{-3}$, a fixed learning rate of $3 \times 10^{-3}$, and a batch size of $128$.
We evaluate each probe's performance on an additional 10k episodes.

\subsection{LeWM Benchmarks}
For experiments on \textbf{PushT}, \textbf{Two-Room}, \textbf{OG-Cube}, and \textbf{Reacher}, we build on top of the original implementation provided by \cite{LeWorldModel} together with the provided expert rollouts for each task.
We use a history size of three for all four tasks. As in \cite{LeWorldModel}, during episode sampling, we apply a 5-frame skip and group the consecutive actions between sampled frames into a single action vector.


\begin{table*}[h!]
    \centering
    \small
    \begin{tabular}{@{}llccc@{}}
    \toprule
    & & \multicolumn{3}{c}{Selected}\\
    \cmidrule(l){3-5}
    Method & Search space & Pong & Dino & Golf\\
    \midrule
    SMWM* & $\lambda_{\mathrm{IDM}} \in \{0.1, 0.2, 0.5, 1, 2, 5, 10, 20, 50, 100\}$ & $100$ & $0.2$ & $1$\\
    LeWM* & $\lambda_{\mathrm{SIGReg}} \in \{0.1, 0.2, 0.5, 1, 2, 5, 10, 20, 50, 100\}$ & $100$ & $2$ & $0.1$\\
    LeWM-Time* & $\lambda_{\mathrm{SIGReg}} \in \{0.1, 0.2, 0.5, 1, 2, 5, 10, 20, 50, 100\}$ & $10$ & $2$ & $20$\\
    LeWM-Detached* & $\lambda_{\mathrm{SIGReg}} \in \{0.1, 0.2, 0.5, 1, 2, 5, 10, 20, 50, 100\}$ & $0.2$ & $20$ & $0.02$\\
    LeWM-Flat* & $\lambda_{\mathrm{SIGReg}} \in \{0.1, 0.2, 0.5, 1, 2, 5, 10, 20, 50, 100\}$ & $1$ & $0.2$ & $1$\\
    \midrule
    \textbf{\modelname (Ours)} & $\lambda_z{=}0.25$, $\lambda_{\mathrm{pred}}{=}0.5$, $\lambda_d{=}2$ (fixed) & \multicolumn{3}{c}{all environments}\\
    \bottomrule
    \end{tabular}

    \caption{\textbf{Hyperparameter Search Configurations and Selected Parameters}. We show the search space of each method together with the selected hyperparameters for each environment.}
    \label{tab:appendix_hp_grid}
\end{table*}

\begin{figure*}[!ht]
  \centering{\includegraphics[width=0.8\textwidth]{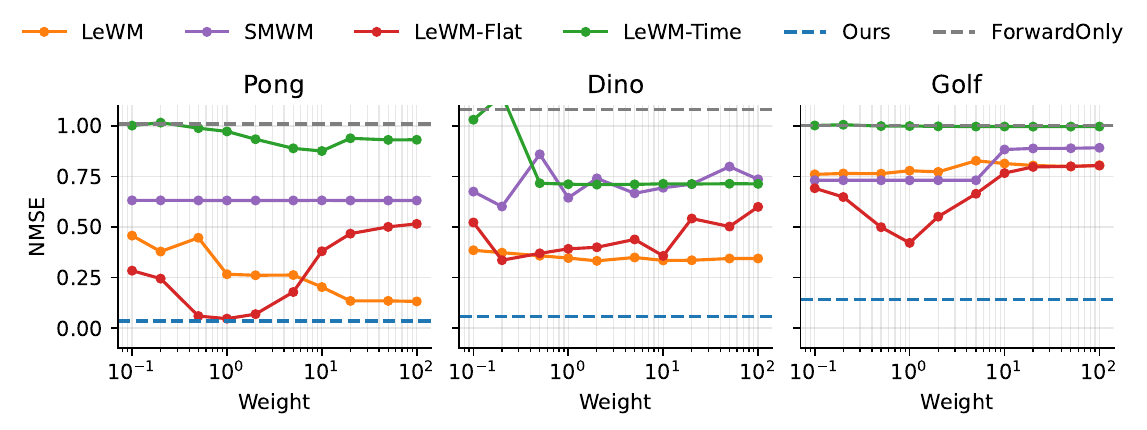}}
  \caption{\textbf{Hyperparameter Search Visualization.} The NMSE loss is estimated from offline residual MLP probes on the predicted 5-step rollout $\hat{z}$.}
  \label{fig:appendix_hp_grid_search}
\end{figure*}
\section{Hyperparameter Search}     
For baselines trained on the game environments, we perform a grid search over full 10M episode runs.
For SMWM~\cite{SMWM}, we grid-search the action-prediction loss weight; for LeWM~\cite{LeWorldModel}, LeWM-Detached, LeWM-Time, and LeWM-Flat, we grid-search the SIGReg weight.
The selected configuration is the final run with the lowest NMSE of an offline 5-step Residual MLP probe, averaged over all probing-table state groups.
We use the probe of the predicted rollout $\hat{z}$ instead of the direct image embedding $z$, as we are interested in a world model that is able to predict future states accurately, not a pure image embedding.
Results of the hyperparameter search and sensitivity are shown in \cref{tab:appendix_hp_grid} and \cref{fig:appendix_hp_grid_search}.

\section{Measuring Latent Path Straightness}
For each environment, we estimate the path straightness shown in \cref{tab:path_straightness} by sampling $N=100$ pairs of start and end states from the state variables defined in \cref{sec:appendix_game_environments}, with two exceptions: the score in Pong and the wood texture offset in Golf.
We exclude the Pong score because its 0 to 99 range introduces non-continuous, looping visual features as the smaller digit rolls over.
Similarly, the Golf wood texture offset is omitted because it produces texture crops that lack a meaningful linear structure.
We then compute $T=64$ linearly interpolated states for each trajectory.
These states are rendered as images and encoded with the respective image encoder $\text{Enc}_\theta$, yielding a tensor of latent trajectories $z \in \mathbb{R}^{N \times T \times 192}$.
Finally, we compute the path straightness as the average ratio of the arc length to the chord length:
\begin{align}
    \frac{1}{N} \sum_{i=1}^N \frac{\sum_{t=1}^{T-1} \lVert z_{i, t+1} - z_{i, t} \rVert_2}{\lVert z_{i, T} - z_{i, 1} \rVert_2}
\end{align}
We use this ratio because it is less sensitive to the choice of interpolation step size than cosine-based curvature, which depends on step granularity and is especially noisy for the small, low-resolution images in our setting.

\begin{table*}[h!]
    \centering
    \footnotesize
    \begin{subtable}[t]{\linewidth}
    \centering
    \caption{Offline Linear Probe NMSE$\downarrow$ of Image Embedding $z$}
    \begin{tabular}{@{}lc@{\hskip 4pt}c@{\hskip 4pt}c@{\hskip 4pt}cc@{\hskip 4pt}c@{\hskip 4pt}c@{\hskip 4pt}c@{\hskip 4pt}c@{\hskip 4pt}cc@{\hskip 4pt}c@{\hskip 4pt}c@{\hskip 4pt}c@{\hskip 4pt}c@{}}
    \toprule
    & \multicolumn{4}{c}{Pong} & \multicolumn{6}{c}{Dino} & \multicolumn{5}{c}{Golf}\\
    Method & avg. & ball & bats & score & avg. & cactus & clouds & dino & hearts & sky & avg. & bars & cross & ball & wood\\
    \midrule
    ForwardOnly & $0.965$ & $0.999$ & $1.000$ & $0.899$ & $0.911$ & $1.002$ & $0.999$ & $1.001$ & $1.001$ & $0.239$ & $1.002$ & $1.002$ & $1.001$ & $1.002$ & $1.004$\\
    Random & $0.507$ & $0.607$ & $0.390$ & $0.532$ & $0.536$ & $0.931$ & $0.717$ & $0.484$ & $0.089$ & $\mathbf{0.002}$ & $0.917$ & $0.973$ & $0.808$ & $1.003$ & $0.788$\\
    \midrule
    SMWM~\cite{SMWM} & $0.662$ & $1.008$ & $\mathbf{0.000}$ & $1.013$ & $0.655$ & $\underline{0.643}$ & $1.003$ & $0.044$ & $0.887$ & $0.099$ & $0.748$ & $1.027$ & $1.027$ & $\mathbf{0.000}$ & $1.031$\\
    LeWM~\cite{LeWorldModel} & $0.722$ & $1.087$ & $1.086$ & $\underline{0.025}$ & $0.558$ & $1.033$ & $0.502$ & $1.045$ & $0.030$ & $0.051$ & $0.975$ & $1.086$ & $1.064$ & $1.005$ & $\mathbf{0.727}$\\
    LeWM-Time & $0.993$ & $0.994$ & $0.995$ & $0.989$ & $0.904$ & $1.000$ & $0.993$ & $0.999$ & $0.993$ & $0.205$ & $1.000$ & $1.000$ & $1.000$ & $1.000$ & $0.997$\\
    LeWM-Detached & $0.411$ & $0.706$ & $0.429$ & $0.126$ & $0.762$ & $1.082$ & $0.858$ & $1.054$ & $0.149$ & $0.204$ & $0.985$ & $1.020$ & $0.778$ & $1.035$ & $0.979$\\
    LeWM-Flat & $0.391$ & $1.069$ & $0.120$ & $0.048$ & $\underline{0.445}$ & $0.785$ & $0.767$ & $\mathbf{0.021}$ & $\underline{0.026}$ & $0.012$ & $0.947$ & $1.090$ & $0.542$ & $1.069$ & $0.797$\\
    \midrule
    SMWM* & $0.651$ & $0.945$ & $\underline{0.001}$ & $1.037$ & $0.665$ & $0.703$ & $1.000$ & $0.041$ & $0.956$ & $0.083$ & $0.748$ & $1.027$ & $1.027$ & $\underline{0.000}$ & $1.031$\\
    LeWM* & $0.221$ & $0.387$ & $0.111$ & $0.183$ & $0.498$ & $1.082$ & $\underline{0.362}$ & $1.051$ & $0.043$ & $0.029$ & $0.973$ & $1.085$ & $1.074$ & $0.972$ & $0.756$\\
    LeWM-Time* & $1.014$ & $1.002$ & $1.015$ & $1.024$ & $0.969$ & $0.790$ & $1.018$ & $0.900$ & $1.003$ & $1.013$ & $1.011$ & $1.004$ & $0.986$ & $1.022$ & $1.020$\\
    LeWM-Detached* & $0.251$ & $0.538$ & $0.105$ & $0.137$ & $0.539$ & $1.083$ & $0.422$ & $1.049$ & $0.089$ & $0.096$ & $0.982$ & $1.090$ & $1.079$ & $1.003$ & $0.749$\\
    LeWM-Flat* & $\underline{0.145}$ & $\underline{0.276}$ & $0.077$ & $0.096$ & $0.454$ & $0.927$ & $0.742$ & $\underline{0.030}$ & $0.041$ & $0.031$ & $\underline{0.559}$ & $\underline{0.534}$ & $\mathbf{0.202}$ & $0.496$ & $0.846$\\
    \midrule
    \textbf{\modelname (Ours)} & $\mathbf{0.038}$ & $\mathbf{0.084}$ & $0.010$ & $\mathbf{0.023}$ & $\mathbf{0.174}$ & $\mathbf{0.416}$ & $\mathbf{0.242}$ & $0.080$ & $\mathbf{0.020}$ & $\underline{0.011}$ & $\mathbf{0.407}$ & $\mathbf{0.337}$ & $\underline{0.422}$ & $0.195$ & $\underline{0.746}$\\
    \bottomrule
    \end{tabular}
    \end{subtable}
    \vspace{0.5em}

    \begin{subtable}[t]{\linewidth}
    \centering
    \caption{Offline Linear Probe NMSE$\downarrow$ of Predicted Rollout $\hat{z}$}
    \begin{tabular}{@{}lc@{\hskip 4pt}c@{\hskip 4pt}c@{\hskip 4pt}cc@{\hskip 4pt}c@{\hskip 4pt}c@{\hskip 4pt}c@{\hskip 4pt}c@{\hskip 4pt}cc@{\hskip 4pt}c@{\hskip 4pt}c@{\hskip 4pt}c@{\hskip 4pt}c@{}}
    \toprule
    & \multicolumn{4}{c}{Pong} & \multicolumn{6}{c}{Dino} & \multicolumn{5}{c}{Golf}\\
    Method & avg. & ball & bats & score & avg. & cactus & clouds & dino & hearts & sky & avg. & bars & cross & ball & wood\\
    \midrule
    ForwardOnly & $1.000$ & $1.002$ & $1.001$ & $0.997$ & $0.999$ & $1.001$ & $1.002$ & $1.000$ & $1.003$ & $0.979$ & $1.002$ & $1.002$ & $1.001$ & $1.000$ & $1.004$\\
    Random & $0.877$ & $0.913$ & $0.837$ & $0.891$ & $0.808$ & $1.024$ & $0.937$ & $0.928$ & $0.607$ & $0.093$ & $0.988$ & $1.027$ & $0.986$ & $1.008$ & $0.910$\\
    \midrule
    SMWM~\cite{SMWM} & $0.638$ & $1.015$ & $\mathbf{0.000}$ & $1.014$ & $0.721$ & $0.907$ & $1.003$ & $0.188$ & $0.950$ & $0.134$ & $0.749$ & $1.029$ & $1.021$ & $\mathbf{0.005}$ & $1.035$\\
    LeWM~\cite{LeWorldModel} & $0.706$ & $1.091$ & $1.083$ & $\mathbf{0.026}$ & $0.564$ & $1.065$ & $0.503$ & $1.049$ & $0.038$ & $0.055$ & $0.981$ & $1.086$ & $1.069$ & $1.022$ & $\mathbf{0.733}$\\
    LeWM-Time & $0.996$ & $1.000$ & $0.989$ & $1.000$ & $0.994$ & $1.000$ & $1.001$ & $1.000$ & $1.001$ & $0.944$ & $1.000$ & $1.000$ & $1.000$ & $0.999$ & $1.001$\\
    LeWM-Detached & $0.418$ & $0.778$ & $0.436$ & $0.129$ & $0.764$ & $1.065$ & $0.861$ & $1.046$ & $0.170$ & $0.215$ & $1.005$ & $1.048$ & $0.894$ & $1.025$ & $0.973$\\
    LeWM-Flat & $0.365$ & $1.084$ & $0.138$ & $0.058$ & $0.477$ & $0.978$ & $0.772$ & $\mathbf{0.069}$ & $\underline{0.030}$ & $\underline{0.018}$ & $0.961$ & $1.085$ & $0.668$ & $1.071$ & $0.795$\\
    \midrule
    SMWM* & $0.653$ & $1.037$ & $\underline{0.001}$ & $1.039$ & $0.721$ & $0.944$ & $1.002$ & $0.160$ & $0.981$ & $0.122$ & $0.749$ & $1.029$ & $1.021$ & $\underline{0.005}$ & $1.035$\\
    LeWM* & $0.262$ & $0.623$ & $0.095$ & $0.162$ & $0.492$ & $1.059$ & $\underline{0.352}$ & $1.047$ & $0.056$ & $0.028$ & $0.977$ & $1.085$ & $1.067$ & $0.984$ & $0.764$\\
    LeWM-Time* & $1.003$ & $0.959$ & $1.022$ & $1.017$ & $0.947$ & $\underline{0.786}$ & $1.009$ & $0.832$ & $0.984$ & $1.003$ & $1.011$ & $1.011$ & $1.010$ & $1.010$ & $1.013$\\
    LeWM-Detached* & $0.274$ & $0.682$ & $0.110$ & $0.137$ & $0.526$ & $1.057$ & $0.400$ & $1.043$ & $0.104$ & $0.085$ & $0.984$ & $1.084$ & $1.066$ & $1.020$ & $0.751$\\
    LeWM-Flat* & $\underline{0.110}$ & $\underline{0.294}$ & $0.027$ & $0.058$ & $\underline{0.471}$ & $0.987$ & $0.742$ & $\underline{0.086}$ & $0.044$ & $0.038$ & $\underline{0.636}$ & $\underline{0.691}$ & $\mathbf{0.387}$ & $0.492$ & $0.838$\\
    \midrule
    \textbf{\modelname (Ours)} & $\mathbf{0.069}$ & $\mathbf{0.208}$ & $0.010$ & $\underline{0.026}$ & $\mathbf{0.191}$ & $\mathbf{0.492}$ & $\mathbf{0.239}$ & $0.130$ & $\mathbf{0.024}$ & $\mathbf{0.012}$ & $\mathbf{0.465}$ & $\mathbf{0.444}$ & $\underline{0.525}$ & $0.217$ & $\underline{0.745}$\\
    \bottomrule
    \end{tabular}
    \end{subtable}

    \caption{\textbf{Offline linear probes}. Our method achieves the lowest average feature probing loss across all environments, indicating not only good feature coverage but also better linear separability.}
    \label{tab:appendix_games_linear_offline_probes}
\end{table*}

\section{State Probing on LeWorldModel Benchmarks with Static Wood-Texture Backgrounds}
\label{sec:appendix_lewm_probing}
We evaluate the task-relevant physical quantities used for controlling each environment with offline MLP state probes.
For \textbf{PushT}, we probe the 2D position of the agent, the 2D position of the T-shaped object, and the object orientation.
For \textbf{TwoRoom}, we probe the 2D agent position.
For \textbf{OG-Cube}, we probe the 3D positions and yaw angles of both the robot end-effector and the manipulated object.
For \textbf{Reacher}, we probe the two robot joint angles, the 2D fingertip position, and the 2D target position.
For translational variables, we report the mean Euclidean distance in the native coordinates of the corresponding environment.
For angular quantities, we report the mean absolute angular error in radians.
Lower values are better in all cases.
We compare LeWorldModel, LeWM-Flat, IDM, our method, and our method combined with IDM.
For LeWorldModel, we use the original loss weights $\lambda_{\mathrm{pred}}=1$ and $\lambda_{\mathrm{sigreg}}=0.09$~\cite{LeWorldModel}.
LeWM-Flat applies SigReg after flattening the batch and temporal dimensions and uses $\lambda_{\mathrm{sigreg}}=1.0$, selected from the hyperparameter sweep on the game environments above.
For IDM, we use $\lambda_{\mathrm{pred}}=1$ and $\lambda_{\mathrm{IDM}}=1$, as our hyperparameter search on the game environments above showed little sensitivity to the weighting.
For our method, we use $\lambda_z=0.25$, $\lambda_{\mathrm{pred}}=0.5$, and $\lambda_{d}=2$.

\begin{table*}[t]
    \centering
    \small
    \setlength{\tabcolsep}{4.0pt}
    \begin{tabular}{llccc c}
        \toprule
        & & \multicolumn{3}{c}{PushT} & TwoRoom \\
        \cmidrule(lr){3-5}\cmidrule(lr){6-6}
        Method & Probe & Agent pos. $\downarrow$ & Object pos. $\downarrow$ & Object angle $\downarrow$ & Agent pos. $\downarrow$ \\
        \midrule
        LeWM & Linear & 131.61 & 87.03 & 0.859 & 47.01 \\
        LeWM & MLP & 121.17 & 73.66 & 0.788 & 45.33 \\
        LeWM-Flat & Linear & 81.80 & 50.63 & 0.754 & 39.44 \\
        LeWM-Flat & MLP & 55.67 & 28.88 & 0.419 & 37.01 \\
        \midrule
        IDM & Linear & 3.99 & \textbf{24.47} & \textbf{0.670} & 3.71 \\
        IDM & MLP & \textbf{3.66} & 18.01 & 0.413 & 3.29 \\
        \midrule
        Ours & Linear & 17.46 & 27.33 & 0.763 & 6.34 \\
        Ours & MLP & 8.60 & 18.20 & 0.397 & 1.48 \\
        \midrule
        Ours+IDM & Linear & \textbf{3.58} & 26.52 & 0.726 & \textbf{1.55} \\
        Ours+IDM & MLP & 4.32 & \textbf{17.12} & \textbf{0.367} & \textbf{0.73} \\
        \bottomrule
    \end{tabular}
    \caption{Physical-state probing on \textbf{PushT} and \textbf{TwoRoom} with a static wood-texture distractor. Translational quantities are reported as mean Euclidean error in the native environment coordinates, while angular error is the mean absolute angular error in radians. Lower is better, with the best MLP and Linear probe results in \textbf{bold}.}
    \label{tab:probe_pusht_tworoom}
\end{table*}

\begin{table*}[t]
    \centering
    \small
    \setlength{\tabcolsep}{4.0pt}
    \begin{tabular}{llcccc cccc}
        \toprule
        & & \multicolumn{4}{c}{OG-Cube} & \multicolumn{4}{c}{Reacher} \\
        \cmidrule(lr){3-6}\cmidrule(lr){7-10}
        Method & Probe & Agent xyz & Object xyz & Agent yaw & Object yaw & Joint 0 & Joint 1 & Fingertip xy & Target xy \\
        \midrule
        LeWM & Linear & 0.182 & 0.187 & 1.394 & \textbf{1.436} & 1.468 & 1.378 & 0.169 & \textbf{0.124} \\
        LeWM & MLP & 0.130 & 0.126 & \textbf{0.588} & \textbf{0.529} & 0.735 & 0.880 & 0.126 & 0.052 \\
        LeWM-Flat & Linear & 0.132 & 0.151 & 1.417 & 1.508 & 0.292 & 1.078 & 0.126 & 0.125 \\
        LeWM-Flat & MLP & 0.0162 & 0.0546 & 1.325 & 1.382 & 0.111 & 0.553 & 0.0738 & 0.121 \\
        \midrule
        IDM & Linear & \textbf{0.0030} & \textbf{0.0068} & \textbf{1.371} & 1.539 & 1.385 & 0.895 & 0.168 & 0.125 \\
        IDM & MLP & \textbf{0.0026} & \textbf{0.0046} & 1.274 & 1.522 & 1.162 & 0.820 & 0.166 & 0.125 \\
        \midrule
        Ours & Linear & 0.0236 & 0.0214 & 1.403 & 1.473 & 0.0265 & 0.1246 & 0.0241 & 0.124 \\
        Ours & MLP & 0.0113 & 0.0103 & 0.732 & 0.708 & 0.0141 & 0.0321 & 0.0061 & \textbf{0.0463} \\
        \midrule
        Ours+IDM & Linear & 0.0040 & 0.0089 & 1.374 & 1.482 & \textbf{0.0139} & \textbf{0.0392} & \textbf{0.0107} & 0.124 \\
        Ours+IDM & MLP & 0.0048 & 0.0059 & 0.808 & 0.841 & \textbf{0.0100} & \textbf{0.0234} & \textbf{0.0042} & 0.0487 \\
        \bottomrule
    \end{tabular}
    \caption{Physical-state probing on \textbf{OG-Cube} and \textbf{Reacher} with a static wood-texture distractor. Position errors are mean Euclidean errors, while angular quantities are mean absolute errors in radians. Lower is better, with the best MLP and Linear probe results in \textbf{bold}.}
    \label{tab:probe_ogcube_reacher}
\end{table*}

\paragraph{Analysis.}

The probing results reveal substantial differences in the task-relevant physical information retained by the learned representations under static background distractors. LeWM performs poorly across several dynamic state variables, particularly on PushT, TwoRoom, and Reacher. LeWM-Flat improves recoverability in several cases, suggesting that stronger regularization over flattened batch and temporal dimensions partially alleviates the problem, but a substantial gap remains to the dynamics-aware objectives. For example, on PushT the MLP agent-position error decreases from $121.17$ for LeWM to $55.67$ for LeWM-Flat, while our method reaches $8.60$.

Our method substantially improves physical-state recoverability across the evaluated environments. The effect is especially pronounced on Reacher, where the linear probe obtains joint-angle errors of $0.0265$ and $0.1246$,rad and a fingertip-position error of $0.0241$, compared with $1.468$, $1.378$,rad, and $0.169$ for LeWM. The strong linear performance indicates that the controllable robot state is not only retained but also organized in a readily accessible form. IDM shows complementary behavior, providing particularly strong translational probing on PushT and OG-Cube, while remaining substantially weaker than our method on Reacher joint and fingertip states.

Combining our objective with IDM further improves several controllable variables, including TwoRoom agent position and Reacher joint and fingertip states. Overall, the results support the interpretation that static-background degradation is associated with the representation emphasizing highly predictable visual content at the expense of task-relevant dynamics. Stronger SigReg alone, as in LeWM-Flat, helps but does not fully resolve this issue, while our objective and IDM provide stronger supervision for preserving dynamic physical state. Their combination further improves recoverability for several variables, consistent with the complementary planning gains observed in the main experiments.

\subsection{Reconstruction-Based Visualization of Latent Information}
\label{sec:appendix_probe_reconstruction}

\begin{figure*}[h]
    \centering

    \begin{minipage}[t]{0.9\textwidth}
        \centering
        \includegraphics[width=\linewidth]{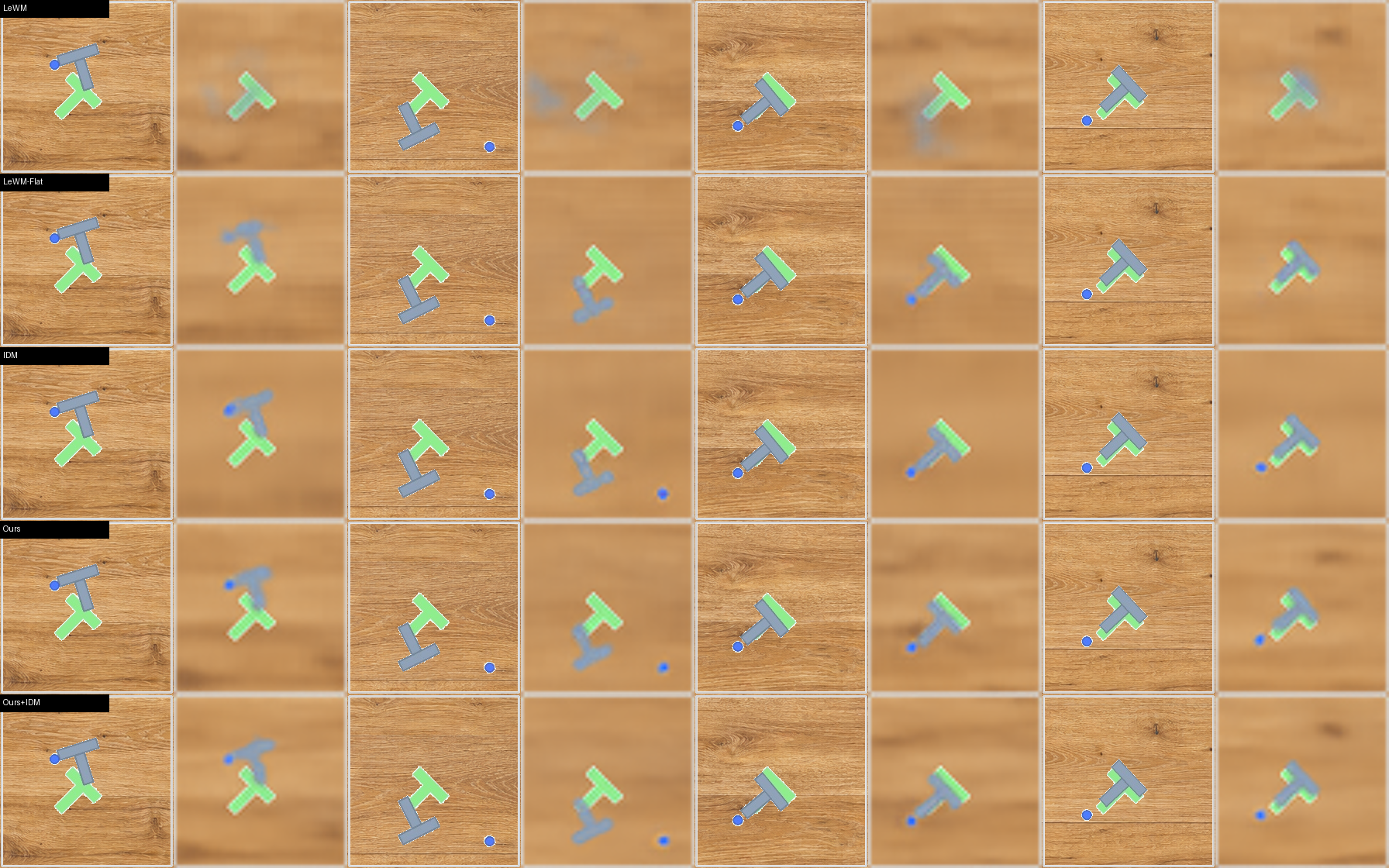}\\[-1mm]
        {\small (a) PushT}
    \end{minipage}

    \vspace{2mm}

    \begin{minipage}[t]{0.9\textwidth}
        \centering
        \includegraphics[width=\linewidth]{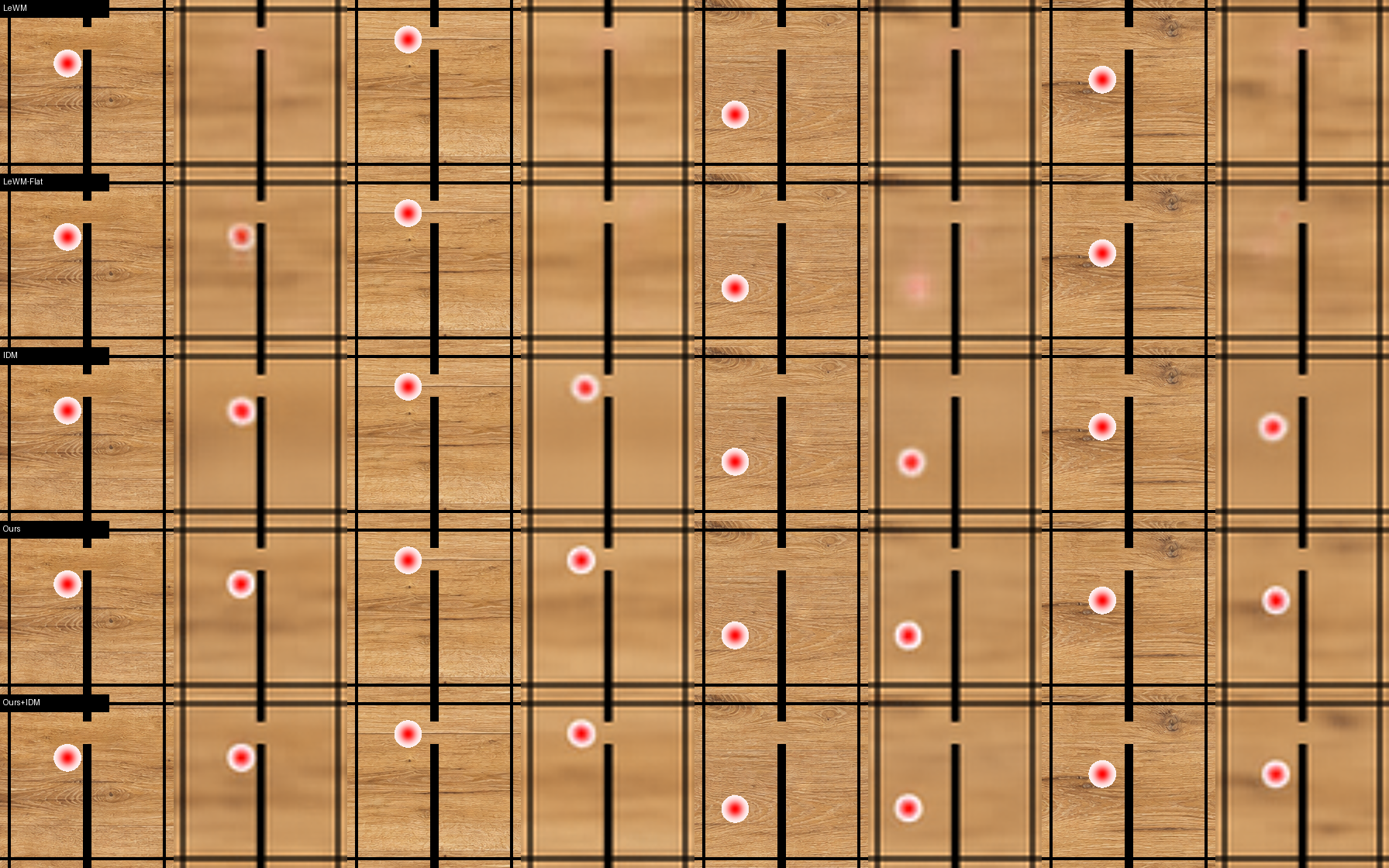}\\[-1mm]
        {\small (b) TwoRoom}
    \end{minipage}

    \caption{\textbf{Qualitative reconstruction from frozen latent representations on PushT and TwoRoom.} Rows within each environment correspond to LeWM, LeWM-Flat, IDM, our method, and Ours+IDM, with original observations and their reconstructions shown alternately. LeWM often reconstructs the dominant static appearance while only weakly recovering the task-relevant foreground. LeWM-Flat improves foreground recovery, while IDM, our method, and Ours+IDM generally preserve the moving entities more clearly.}
    \label{fig:probe_reconstruction_pusht_tworoom}
\end{figure*}

\begin{figure*}[h]
    \centering

    \begin{minipage}[t]{0.9\textwidth}
        \centering
        \includegraphics[width=\linewidth]{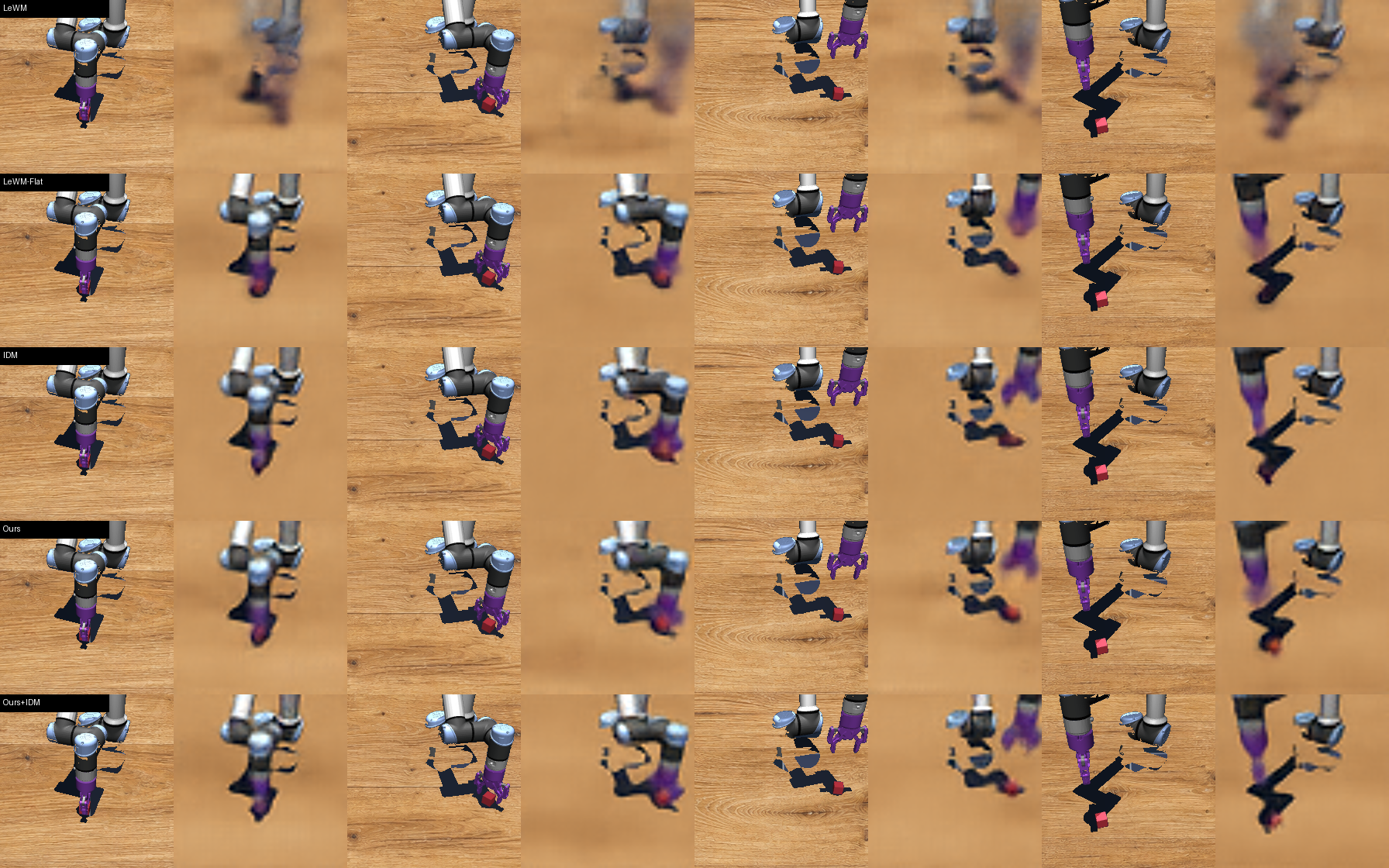}\\[-1mm]
        {\small (a) OG-Cube}
    \end{minipage}

    \vspace{2mm}

    \begin{minipage}[t]{0.9\textwidth}
        \centering
        \includegraphics[width=\linewidth]{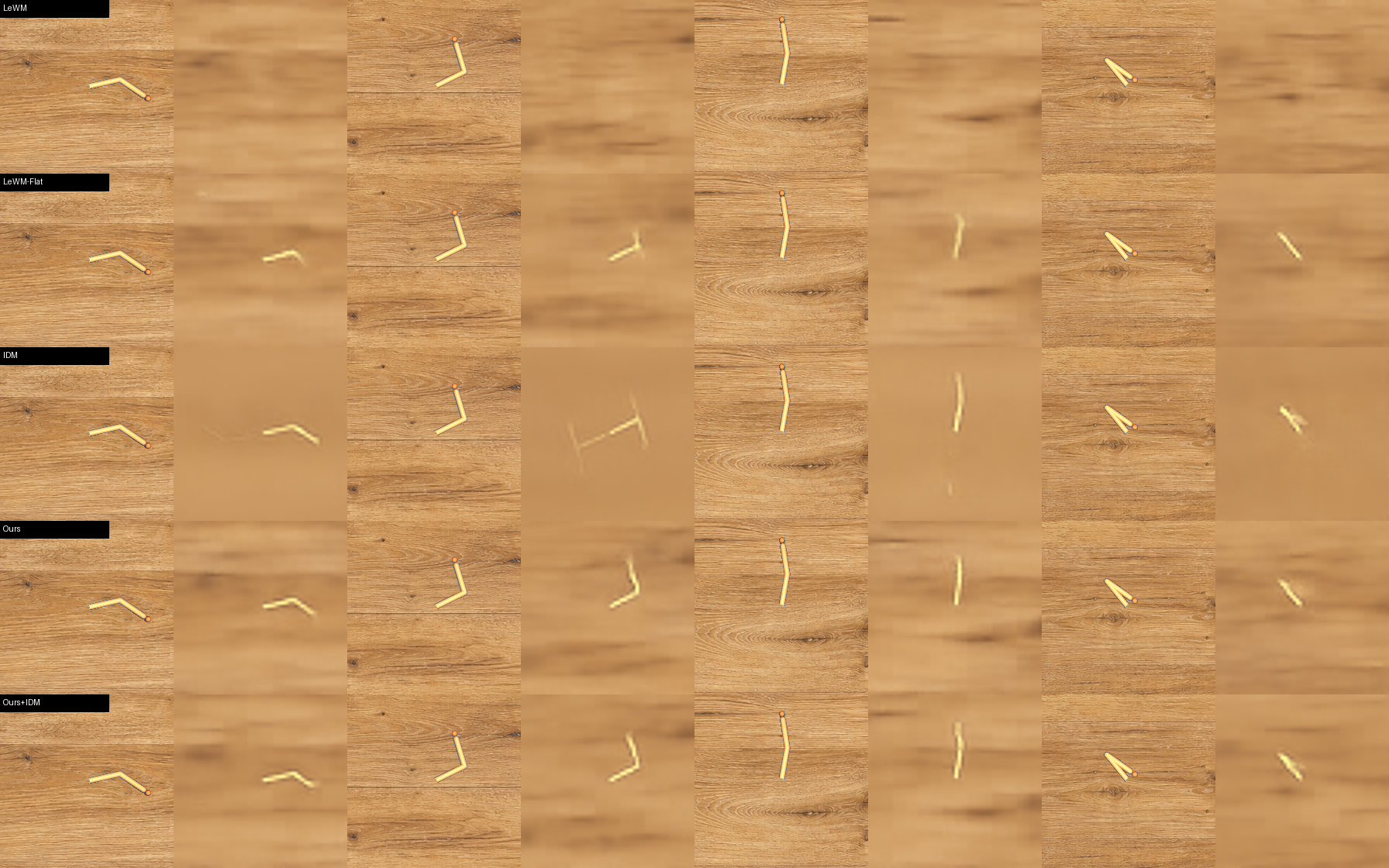}\\[-1mm]
        {\small (b) Reacher}
    \end{minipage}

    \caption{\textbf{Qualitative reconstruction from frozen latent representations on OG-Cube and Reacher.} Rows within each environment correspond to LeWM, LeWM-Flat, IDM, our method, and Ours+IDM, with original observations and their reconstructions shown alternately. Dynamics-aware representations recover the articulated foreground state substantially more clearly than LeWM, while LeWM-Flat provides an intermediate improvement. This is particularly visible on Reacher, where the robot configuration is more faithfully reconstructed despite the dominant wood-texture background.}
    \label{fig:probe_reconstruction_ogcube_reacher}
\end{figure*}

To complement the quantitative probing results, we train a lightweight image decoder on top of each frozen latent representation as an offline visual probe. Figures~\ref{fig:probe_reconstruction_pusht_tworoom} and~\ref{fig:probe_reconstruction_ogcube_reacher} show representative reconstructions for all four environments. Within each row, original observations and their latent reconstructions are shown alternately, with methods ordered as LeWM, LeWM-Flat, IDM, ours, and Ours+IDM.

The qualitative results are consistent with the state probing results. LeWM often reconstructs the dominant static appearance while only weakly recovering the task-relevant foreground state. LeWM-Flat improves foreground reconstruction in several environments, indicating that stronger regularization already alleviates part of this effect, but the recovered dynamic state remains less consistent than for the dynamics-aware objectives. IDM, Ours, and Ours+IDM generally preserve moving agents, manipulated objects, and articulated robot configurations more clearly. This difference is particularly visible on TwoRoom and OG-Cube, where LeWM reconstructions largely lose the foreground configuration, and on Reacher, where the dynamics-aware representations recover the robot pose substantially more clearly.

At the same time, our reconstructions still retain coarse scene appearance rather than becoming invariant to static visual content. Fine wood-texture details are largely smoothed out, while broad background structure remains recoverable. This is desirable because static or slowly varying scene information may still encode useful context such as goals or obstacles. Together with the quantitative probes, these results suggest that the proposed objective reduces the dominance of highly predictable static appearance while preserving both task-relevant dynamics and coarse contextual information.

\begin{table*}[h!]
    \centering
    \small
    \begin{subtable}[t]{\linewidth}
    \centering
    \caption{Offline Residual MLP Probe NMSE$\downarrow$ of Image Embedding $z$}
    \begin{tabular}{@{}lc@{\hskip 4pt}c@{\hskip 4pt}c@{\hskip 4pt}c@{}}
    \toprule
    & \multicolumn{4}{c}{Pong}\\
    Method & avg. & ball & bats & score\\
    \midrule
    ForwardOnly & $1.058 \pm 0.099$ & $\textcolor{red!70!black}{1.260 \pm 0.449}$ & $\textcolor{red!70!black}{1.001 \pm 0.001}$ & $\textcolor{red!70!black}{0.930 \pm 0.121}$\\
    Random & $0.116 \pm 0.009$ & $\textcolor{red!70!black}{0.143 \pm 0.014}$ & $\textcolor{green!60!black}{0.093 \pm 0.011}$ & $\textcolor{red!70!black}{0.115 \pm 0.005}$\\
    \midrule
    SMWM~\cite{SMWM} & $0.657 \pm 0.004$ & $\textcolor{red!70!black}{1.001 \pm 0.001}$ & $\textcolor{green!60!black}{\underline{0.002 \pm 0.000}}$ & $\textcolor{red!70!black}{1.002 \pm 0.001}$\\
    LeWM~\cite{LeWorldModel} & $0.658 \pm 0.003$ & $\textcolor{red!70!black}{0.999 \pm 0.004}$ & $\textcolor{red!70!black}{1.002 \pm 0.000}$ & $\textcolor{green!60!black}{\mathbf{0.003 \pm 0.001}}$\\
    LeWM-Time & $1.077 \pm 0.096$ & $\textcolor{red!70!black}{1.100 \pm 0.145}$ & $\textcolor{red!70!black}{1.071 \pm 0.096}$ & $\textcolor{red!70!black}{1.063 \pm 0.066}$\\
    LeWM-Detached & $0.171 \pm 0.092$ & $\textcolor{red!70!black}{0.174 \pm 0.013}$ & $\textcolor{red!70!black}{0.331 \pm 0.255}$ & $\textcolor{green!60!black}{0.007 \pm 0.000}$\\
    LeWM-Flat & $0.284 \pm 0.006$ & $\textcolor{red!70!black}{0.890 \pm 0.003}$ & $\textcolor{green!60!black}{0.014 \pm 0.011}$ & $\textcolor{green!60!black}{0.005 \pm 0.000}$\\
    \midrule
    SMWM* & $0.651 \pm 0.007$ & $\textcolor{red!70!black}{0.983 \pm 0.029}$ & $\textcolor{green!60!black}{\mathbf{0.002 \pm 0.001}}$ & $\textcolor{red!70!black}{1.003 \pm 0.001}$\\
    LeWM* & $0.061 \pm 0.008$ & $\textcolor{red!70!black}{0.163 \pm 0.025}$ & $\textcolor{green!60!black}{0.018 \pm 0.002}$ & $\textcolor{green!60!black}{0.012 \pm 0.000}$\\
    LeWM-Time* & $0.687 \pm 0.006$ & $\textcolor{green!60!black}{0.022 \pm 0.004}$ & $\textcolor{red!70!black}{0.972 \pm 0.017}$ & $\textcolor{red!70!black}{1.002 \pm 0.001}$\\
    LeWM-Detached* & $0.048 \pm 0.003$ & $\textcolor{red!70!black}{0.103 \pm 0.008}$ & $\textcolor{green!60!black}{0.038 \pm 0.002}$ & $\textcolor{green!60!black}{0.008 \pm 0.001}$\\
    LeWM-Flat* & $\underline{0.008 \pm 0.001}$ & $\textcolor{green!60!black}{\underline{0.012 \pm 0.001}}$ & $\textcolor{green!60!black}{0.004 \pm 0.001}$ & $\textcolor{green!60!black}{0.008 \pm 0.001}$\\
    \midrule
    \textbf{\modelname (Ours)} & $\mathbf{0.004 \pm 0.000}$ & $\textcolor{green!60!black}{\mathbf{0.005 \pm 0.001}}$ & $\textcolor{green!60!black}{0.003 \pm 0.001}$ & $\textcolor{green!60!black}{\underline{0.004 \pm 0.001}}$\\
    \bottomrule
    \end{tabular}
    \end{subtable}
    \vspace{0.5em}

    \begin{subtable}[t]{\linewidth}
    \centering
    \caption{Offline Residual MLP Probe NMSE$\downarrow$ of Predicted Rollout $\hat{z}$}
    \begin{tabular}{@{}lc@{\hskip 4pt}c@{\hskip 4pt}c@{\hskip 4pt}c@{}}
    \toprule
    & \multicolumn{4}{c}{Pong}\\
    Method & avg. & ball & bats & score\\
    \midrule
    ForwardOnly & $1.010 \pm 0.015$ & $1.041 \pm 0.068$ & $1.006 \pm 0.008$ & $0.992 \pm 0.017$\\
    Random & $0.554 \pm 0.022$ & $0.721 \pm 0.031$ & $0.478 \pm 0.028$ & $0.507 \pm 0.019$\\
    \midrule
    SMWM~\cite{SMWM} & $0.630 \pm 0.003$ & $1.001 \pm 0.001$ & $\underline{0.002 \pm 0.000}$ & $1.002 \pm 0.001$\\
    LeWM~\cite{LeWorldModel} & $0.644 \pm 0.003$ & $1.002 \pm 0.000$ & $1.002 \pm 0.001$ & $\mathbf{0.002 \pm 0.000}$\\
    LeWM-Time & $1.001 \pm 0.008$ & $1.003 \pm 0.002$ & $0.998 \pm 0.019$ & $1.002 \pm 0.002$\\
    LeWM-Detached & $0.239 \pm 0.095$ & $0.397 \pm 0.005$ & $0.347 \pm 0.258$ & $0.007 \pm 0.001$\\
    LeWM-Flat & $0.278 \pm 0.013$ & $0.968 \pm 0.017$ & $0.039 \pm 0.035$ & $0.005 \pm 0.000$\\
    \midrule
    SMWM* & $0.630 \pm 0.002$ & $1.001 \pm 0.001$ & $\mathbf{0.001 \pm 0.000}$ & $1.002 \pm 0.001$\\
    LeWM* & $0.138 \pm 0.005$ & $0.471 \pm 0.019$ & $0.020 \pm 0.000$ & $0.009 \pm 0.001$\\
    LeWM-Time* & $0.868 \pm 0.028$ & $0.507 \pm 0.101$ & $1.002 \pm 0.001$ & $1.002 \pm 0.001$\\
    LeWM-Detached* & $0.129 \pm 0.011$ & $0.395 \pm 0.037$ & $0.053 \pm 0.004$ & $0.008 \pm 0.001$\\
    LeWM-Flat* & $\underline{0.054 \pm 0.003}$ & $\underline{0.183 \pm 0.016}$ & $0.004 \pm 0.001$ & $0.007 \pm 0.002$\\
    \midrule
    \textbf{\modelname (Ours)} & $\mathbf{0.036 \pm 0.001}$ & $\mathbf{0.122 \pm 0.003}$ & $0.004 \pm 0.000$ & $\underline{0.005 \pm 0.001}$\\
    \bottomrule
    \end{tabular}
    \end{subtable}

    \caption{\textbf{Pong Standard Deviation of Three Seed Runs}.}
    \label{tab:appendix_pong_std}
\end{table*}

\begin{table*}[h!]
    \centering
    \small
    \begin{subtable}[t]{\linewidth}
    \centering
    \caption{Offline Residual MLP Probe NMSE$\downarrow$ of Image Embedding $z$}
    \begin{tabular}{@{}lc@{\hskip 4pt}c@{\hskip 4pt}c@{\hskip 4pt}c@{\hskip 4pt}c@{\hskip 4pt}c@{}}
    \toprule
    & \multicolumn{6}{c}{Dino}\\
    Method & avg. & cactus & clouds & dino & hearts & sky\\
    \midrule
    ForwardOnly & $0.949 \pm 0.054$ & $\textcolor{red!70!black}{1.003 \pm 0.004}$ & $\textcolor{red!70!black}{1.002 \pm 0.010}$ & $\textcolor{red!70!black}{1.002 \pm 0.003}$ & $\textcolor{red!70!black}{1.031 \pm 0.045}$ & $\textcolor{red!70!black}{0.511 \pm 0.422}$\\
    Random & $0.361 \pm 0.006$ & $\textcolor{red!70!black}{0.833 \pm 0.010}$ & $\textcolor{red!70!black}{0.535 \pm 0.016}$ & $\textcolor{red!70!black}{0.116 \pm 0.005}$ & $\textcolor{green!60!black}{0.031 \pm 0.002}$ & $\textcolor{green!60!black}{\mathbf{0.004 \pm 0.002}}$\\
    \midrule
    SMWM~\cite{SMWM} & $0.649 \pm 0.040$ & $\textcolor{red!70!black}{0.434 \pm 0.213}$ & $\textcolor{red!70!black}{1.075 \pm 0.061}$ & $\textcolor{green!60!black}{0.028 \pm 0.028}$ & $\textcolor{red!70!black}{0.787 \pm 0.061}$ & $\textcolor{green!60!black}{0.071 \pm 0.057}$\\
    LeWM~\cite{LeWorldModel} & $0.394 \pm 0.032$ & $\textcolor{red!70!black}{0.769 \pm 0.304}$ & $\textcolor{red!70!black}{0.248 \pm 0.073}$ & $\textcolor{red!70!black}{0.983 \pm 0.033}$ & $\textcolor{green!60!black}{0.012 \pm 0.002}$ & $\textcolor{green!60!black}{0.005 \pm 0.002}$\\
    LeWM-Time & $0.913 \pm 0.019$ & $\textcolor{red!70!black}{1.003 \pm 0.001}$ & $\textcolor{red!70!black}{0.992 \pm 0.013}$ & $\textcolor{red!70!black}{1.002 \pm 0.002}$ & $\textcolor{red!70!black}{1.029 \pm 0.034}$ & $\textcolor{red!70!black}{0.248 \pm 0.130}$\\
    LeWM-Detached & $0.645 \pm 0.020$ & $\textcolor{red!70!black}{1.002 \pm 0.002}$ & $\textcolor{red!70!black}{0.727 \pm 0.043}$ & $\textcolor{red!70!black}{1.002 \pm 0.002}$ & $\textcolor{green!60!black}{0.017 \pm 0.007}$ & $\textcolor{green!60!black}{0.006 \pm 0.002}$\\
    LeWM-Flat & $0.331 \pm 0.008$ & $\textcolor{red!70!black}{0.522 \pm 0.073}$ & $\textcolor{red!70!black}{0.593 \pm 0.026}$ & $\textcolor{green!60!black}{\mathbf{0.002 \pm 0.000}}$ & $\textcolor{green!60!black}{\mathbf{0.006 \pm 0.001}}$ & $\textcolor{green!60!black}{0.007 \pm 0.001}$\\
    \midrule
    SMWM* & $0.606 \pm 0.042$ & $\textcolor{red!70!black}{0.398 \pm 0.047}$ & $\textcolor{red!70!black}{1.005 \pm 0.038}$ & $\textcolor{green!60!black}{0.003 \pm 0.000}$ & $\textcolor{red!70!black}{0.775 \pm 0.177}$ & $\textcolor{green!60!black}{0.069 \pm 0.080}$\\
    LeWM* & $\underline{0.282 \pm 0.013}$ & $\textcolor{red!70!black}{0.990 \pm 0.008}$ & $\textcolor{green!60!black}{\underline{0.059 \pm 0.006}}$ & $\textcolor{red!70!black}{0.751 \pm 0.072}$ & $\textcolor{green!60!black}{0.011 \pm 0.002}$ & $\textcolor{green!60!black}{0.012 \pm 0.002}$\\
    LeWM-Time* & $0.697 \pm 0.002$ & $\textcolor{green!60!black}{\mathbf{0.005 \pm 0.001}}$ & $\textcolor{red!70!black}{1.004 \pm 0.002}$ & $\textcolor{green!60!black}{0.003 \pm 0.001}$ & $\textcolor{red!70!black}{0.971 \pm 0.007}$ & $\textcolor{red!70!black}{1.004 \pm 0.002}$\\
    LeWM-Detached* & $0.332 \pm 0.004$ & $\textcolor{red!70!black}{1.005 \pm 0.002}$ & $\textcolor{green!60!black}{0.073 \pm 0.003}$ & $\textcolor{red!70!black}{0.959 \pm 0.011}$ & $\textcolor{green!60!black}{0.011 \pm 0.004}$ & $\textcolor{green!60!black}{0.017 \pm 0.002}$\\
    LeWM-Flat* & $0.301 \pm 0.008$ & $\textcolor{red!70!black}{0.650 \pm 0.128}$ & $\textcolor{red!70!black}{0.498 \pm 0.033}$ & $\textcolor{green!60!black}{\underline{0.002 \pm 0.000}}$ & $\textcolor{green!60!black}{0.007 \pm 0.001}$ & $\textcolor{green!60!black}{0.016 \pm 0.003}$\\
    \midrule
    \textbf{\modelname (Ours)} & $\mathbf{0.021 \pm 0.002}$ & $\textcolor{green!60!black}{\underline{0.011 \pm 0.002}}$ & $\textcolor{green!60!black}{\mathbf{0.037 \pm 0.003}}$ & $\textcolor{green!60!black}{0.006 \pm 0.003}$ & $\textcolor{green!60!black}{\underline{0.006 \pm 0.002}}$ & $\textcolor{green!60!black}{\underline{0.004 \pm 0.001}}$\\
    \bottomrule
    \end{tabular}
    \end{subtable}
    \vspace{0.5em}

    \begin{subtable}[t]{\linewidth}
    \centering
    \caption{Offline Residual MLP Probe NMSE$\downarrow$ of Predicted Rollout $\hat{z}$}
    \begin{tabular}{@{}lc@{\hskip 4pt}c@{\hskip 4pt}c@{\hskip 4pt}c@{\hskip 4pt}c@{\hskip 4pt}c@{}}
    \toprule
    & \multicolumn{6}{c}{Dino}\\
    Method & avg. & cactus & clouds & dino & hearts & sky\\
    \midrule
    ForwardOnly & $1.081 \pm 0.067$ & $1.019 \pm 0.016$ & $1.006 \pm 0.005$ & $1.004 \pm 0.004$ & $1.017 \pm 0.024$ & $1.634 \pm 0.537$\\
    Random & $0.617 \pm 0.032$ & $0.974 \pm 0.008$ & $0.764 \pm 0.008$ & $0.631 \pm 0.042$ & $0.279 \pm 0.212$ & $0.026 \pm 0.012$\\
    \midrule
    SMWM~\cite{SMWM} & $0.686 \pm 0.046$ & $0.839 \pm 0.139$ & $1.002 \pm 0.003$ & $0.092 \pm 0.033$ & $0.933 \pm 0.186$ & $0.076 \pm 0.064$\\
    LeWM~\cite{LeWorldModel} & $0.407 \pm 0.024$ & $0.856 \pm 0.143$ & $0.251 \pm 0.065$ & $0.995 \pm 0.010$ & $0.017 \pm 0.003$ & $\mathbf{0.005 \pm 0.002}$\\
    LeWM-Time & $1.106 \pm 0.103$ & $1.011 \pm 0.009$ & $1.009 \pm 0.005$ & $1.003 \pm 0.002$ & $1.026 \pm 0.036$ & $1.845 \pm 0.860$\\
    LeWM-Detached & $0.646 \pm 0.016$ & $0.988 \pm 0.002$ & $0.732 \pm 0.035$ & $1.002 \pm 0.001$ & $0.028 \pm 0.007$ & $0.006 \pm 0.002$\\
    LeWM-Flat & $0.396 \pm 0.009$ & $0.877 \pm 0.055$ & $0.633 \pm 0.028$ & $0.055 \pm 0.006$ & $\underline{0.009 \pm 0.001}$ & $0.007 \pm 0.001$\\
    \midrule
    SMWM* & $0.671 \pm 0.015$ & $0.801 \pm 0.022$ & $0.999 \pm 0.005$ & $0.075 \pm 0.013$ & $0.878 \pm 0.081$ & $0.079 \pm 0.072$\\
    LeWM* & $\underline{0.334 \pm 0.003}$ & $0.991 \pm 0.002$ & $\underline{0.060 \pm 0.006}$ & $1.003 \pm 0.001$ & $0.018 \pm 0.001$ & $0.012 \pm 0.002$\\
    LeWM-Time* & $0.710 \pm 0.001$ & $\mathbf{0.120 \pm 0.002}$ & $1.004 \pm 0.002$ & $\mathbf{0.020 \pm 0.000}$ & $0.959 \pm 0.009$ & $1.003 \pm 0.002$\\
    LeWM-Detached* & $0.343 \pm 0.004$ & $0.988 \pm 0.005$ & $0.078 \pm 0.008$ & $1.002 \pm 0.002$ & $0.020 \pm 0.003$ & $0.014 \pm 0.003$\\
    LeWM-Flat* & $0.353 \pm 0.021$ & $0.923 \pm 0.027$ & $0.524 \pm 0.056$ & $0.066 \pm 0.007$ & $0.013 \pm 0.003$ & $0.018 \pm 0.004$\\
    \midrule
    \textbf{\modelname (Ours)} & $\mathbf{0.055 \pm 0.001}$ & $\underline{0.243 \pm 0.012}$ & $\mathbf{0.040 \pm 0.001}$ & $\underline{0.051 \pm 0.004}$ & $\mathbf{0.007 \pm 0.001}$ & $\underline{0.005 \pm 0.001}$\\
    \bottomrule
    \end{tabular}
    \end{subtable}

    \caption{\textbf{Dino Standard Deviation of Three Seed Runs}.}
    \label{tab:appendix_dino_std}
\end{table*}

\begin{table*}[h!]
    \centering
    \small
    \begin{subtable}[t]{\linewidth}
    \centering
    \caption{Offline Residual MLP Probe NMSE$\downarrow$ of Image Embedding $z$}
    \begin{tabular}{@{}lc@{\hskip 4pt}c@{\hskip 4pt}c@{\hskip 4pt}c@{\hskip 4pt}c@{}}
    \toprule
    & \multicolumn{5}{c}{Golf}\\
    Method & avg. & bars & cross & ball & wood\\
    \midrule
    ForwardOnly & $1.980 \pm 0.769$ & $\textcolor{red!70!black}{1.233 \pm 0.200}$ & $\textcolor{red!70!black}{1.558 \pm 0.866}$ & $\textcolor{red!70!black}{1.077 \pm 0.108}$ & $\textcolor{red!70!black}{4.324 \pm 2.622}$\\
    Random & $0.776 \pm 0.006$ & $\textcolor{red!70!black}{0.857 \pm 0.011}$ & $\textcolor{red!70!black}{0.296 \pm 0.018}$ & $\textcolor{red!70!black}{0.989 \pm 0.002}$ & $\textcolor{red!70!black}{0.653 \pm 0.002}$\\
    \midrule
    SMWM~\cite{SMWM} & $0.727 \pm 0.002$ & $\textcolor{red!70!black}{1.000 \pm 0.001}$ & $\textcolor{red!70!black}{0.986 \pm 0.001}$ & $\textcolor{green!60!black}{\mathbf{0.002 \pm 0.001}}$ & $\textcolor{red!70!black}{1.003 \pm 0.001}$\\
    LeWM~\cite{LeWorldModel} & $0.763 \pm 0.009$ & $\textcolor{red!70!black}{1.002 \pm 0.001}$ & $\textcolor{red!70!black}{1.001 \pm 0.000}$ & $\textcolor{red!70!black}{0.942 \pm 0.061}$ & $\textcolor{green!60!black}{0.084 \pm 0.033}$\\
    LeWM-Time & $1.053 \pm 0.076$ & $\textcolor{red!70!black}{1.008 \pm 0.005}$ & $\textcolor{red!70!black}{1.014 \pm 0.025}$ & $\textcolor{red!70!black}{1.004 \pm 0.002}$ & $\textcolor{red!70!black}{1.195 \pm 0.302}$\\
    LeWM-Detached & $0.811 \pm 0.100$ & $\textcolor{red!70!black}{0.924 \pm 0.074}$ & $\textcolor{red!70!black}{0.683 \pm 0.532}$ & $\textcolor{red!70!black}{0.930 \pm 0.037}$ & $\textcolor{red!70!black}{0.572 \pm 0.126}$\\
    LeWM-Flat & $0.694 \pm 0.066$ & $\textcolor{red!70!black}{1.002 \pm 0.001}$ & $\textcolor{red!70!black}{0.379 \pm 0.539}$ & $\textcolor{red!70!black}{0.993 \pm 0.013}$ & $\textcolor{green!60!black}{0.052 \pm 0.025}$\\
    \midrule
    SMWM* & $0.727 \pm 0.002$ & $\textcolor{red!70!black}{1.000 \pm 0.001}$ & $\textcolor{red!70!black}{0.986 \pm 0.001}$ & $\textcolor{green!60!black}{\underline{0.002 \pm 0.001}}$ & $\textcolor{red!70!black}{1.003 \pm 0.001}$\\
    LeWM* & $0.760 \pm 0.007$ & $\textcolor{red!70!black}{1.001 \pm 0.001}$ & $\textcolor{red!70!black}{1.002 \pm 0.001}$ & $\textcolor{red!70!black}{0.909 \pm 0.084}$ & $\textcolor{red!70!black}{0.110 \pm 0.063}$\\
    LeWM-Time* & $0.985 \pm 0.005$ & $\textcolor{red!70!black}{0.972 \pm 0.013}$ & $\textcolor{red!70!black}{0.992 \pm 0.015}$ & $\textcolor{red!70!black}{0.984 \pm 0.002}$ & $\textcolor{red!70!black}{1.002 \pm 0.001}$\\
    LeWM-Detached* & $0.745 \pm 0.008$ & $\textcolor{red!70!black}{1.002 \pm 0.001}$ & $\textcolor{red!70!black}{1.002 \pm 0.001}$ & $\textcolor{red!70!black}{0.908 \pm 0.022}$ & $\textcolor{green!60!black}{\underline{0.048 \pm 0.015}}$\\
    LeWM-Flat* & $\underline{0.299 \pm 0.027}$ & $\textcolor{red!70!black}{\underline{0.317 \pm 0.058}}$ & $\textcolor{green!60!black}{\mathbf{0.059 \pm 0.018}}$ & $\textcolor{red!70!black}{0.312 \pm 0.066}$ & $\textcolor{red!70!black}{0.377 \pm 0.044}$\\
    \midrule
    \textbf{\modelname (Ours)} & $\mathbf{0.061 \pm 0.012}$ & $\textcolor{green!60!black}{\mathbf{0.095 \pm 0.012}}$ & $\textcolor{green!60!black}{\underline{0.083 \pm 0.004}}$ & $\textcolor{green!60!black}{0.033 \pm 0.023}$ & $\textcolor{green!60!black}{\mathbf{0.033 \pm 0.008}}$\\
    \bottomrule
    \end{tabular}
    \end{subtable}
    \vspace{0.5em}

    \begin{subtable}[t]{\linewidth}
    \centering
    \caption{Offline Residual MLP Probe NMSE$\downarrow$ of Predicted Rollout $\hat{z}$}
    \begin{tabular}{@{}lc@{\hskip 4pt}c@{\hskip 4pt}c@{\hskip 4pt}c@{\hskip 4pt}c@{}}
    \toprule
    & \multicolumn{5}{c}{Golf}\\
    Method & avg. & bars & cross & ball & wood\\
    \midrule
    ForwardOnly & $1.003 \pm 0.003$ & $1.002 \pm 0.001$ & $1.001 \pm 0.001$ & $1.005 \pm 0.008$ & $1.005 \pm 0.004$\\
    Random & $0.934 \pm 0.004$ & $0.995 \pm 0.001$ & $0.913 \pm 0.007$ & $0.998 \pm 0.002$ & $0.781 \pm 0.014$\\
    \midrule
    SMWM~\cite{SMWM} & $0.729 \pm 0.002$ & $1.001 \pm 0.000$ & $1.001 \pm 0.000$ & $\mathbf{0.006 \pm 0.001}$ & $1.002 \pm 0.001$\\
    LeWM~\cite{LeWorldModel} & $0.767 \pm 0.011$ & $1.002 \pm 0.000$ & $1.001 \pm 0.001$ & $0.947 \pm 0.063$ & $0.094 \pm 0.034$\\
    LeWM-Time & $1.003 \pm 0.004$ & $1.006 \pm 0.005$ & $1.006 \pm 0.008$ & $1.000 \pm 0.002$ & $1.002 \pm 0.002$\\
    LeWM-Detached & $0.855 \pm 0.054$ & $0.967 \pm 0.038$ & $0.821 \pm 0.313$ & $0.930 \pm 0.045$ & $0.618 \pm 0.102$\\
    LeWM-Flat & $0.722 \pm 0.045$ & $1.002 \pm 0.001$ & $0.590 \pm 0.356$ & $0.994 \pm 0.014$ & $0.061 \pm 0.030$\\
    \midrule
    SMWM* & $0.729 \pm 0.002$ & $1.001 \pm 0.000$ & $1.001 \pm 0.000$ & $\underline{0.006 \pm 0.001}$ & $1.002 \pm 0.001$\\
    LeWM* & $0.763 \pm 0.007$ & $1.002 \pm 0.001$ & $1.002 \pm 0.001$ & $0.913 \pm 0.080$ & $0.117 \pm 0.063$\\
    LeWM-Time* & $0.997 \pm 0.000$ & $0.997 \pm 0.002$ & $0.995 \pm 0.010$ & $0.991 \pm 0.001$ & $1.002 \pm 0.002$\\
    LeWM-Detached* & $0.746 \pm 0.008$ & $1.002 \pm 0.001$ & $1.001 \pm 0.001$ & $0.909 \pm 0.020$ & $\underline{0.051 \pm 0.017}$\\
    LeWM-Flat* & $\underline{0.448 \pm 0.030}$ & $\underline{0.540 \pm 0.048}$ & $\mathbf{0.320 \pm 0.044}$ & $0.402 \pm 0.062$ & $0.425 \pm 0.048$\\
    \midrule
    \textbf{\modelname (Ours)} & $\mathbf{0.164 \pm 0.025}$ & $\mathbf{0.262 \pm 0.029}$ & $\underline{0.348 \pm 0.005}$ & $0.065 \pm 0.049$ & $\mathbf{0.037 \pm 0.006}$\\
    \bottomrule
    \end{tabular}
    \end{subtable}

    \caption{\textbf{Golf Standard Deviation of Three Seed Runs}.}
    \label{tab:appendix_golf_std}
\end{table*}

\clearpage

\begin{figure*}
  \centering{\includegraphics[width=\textwidth]{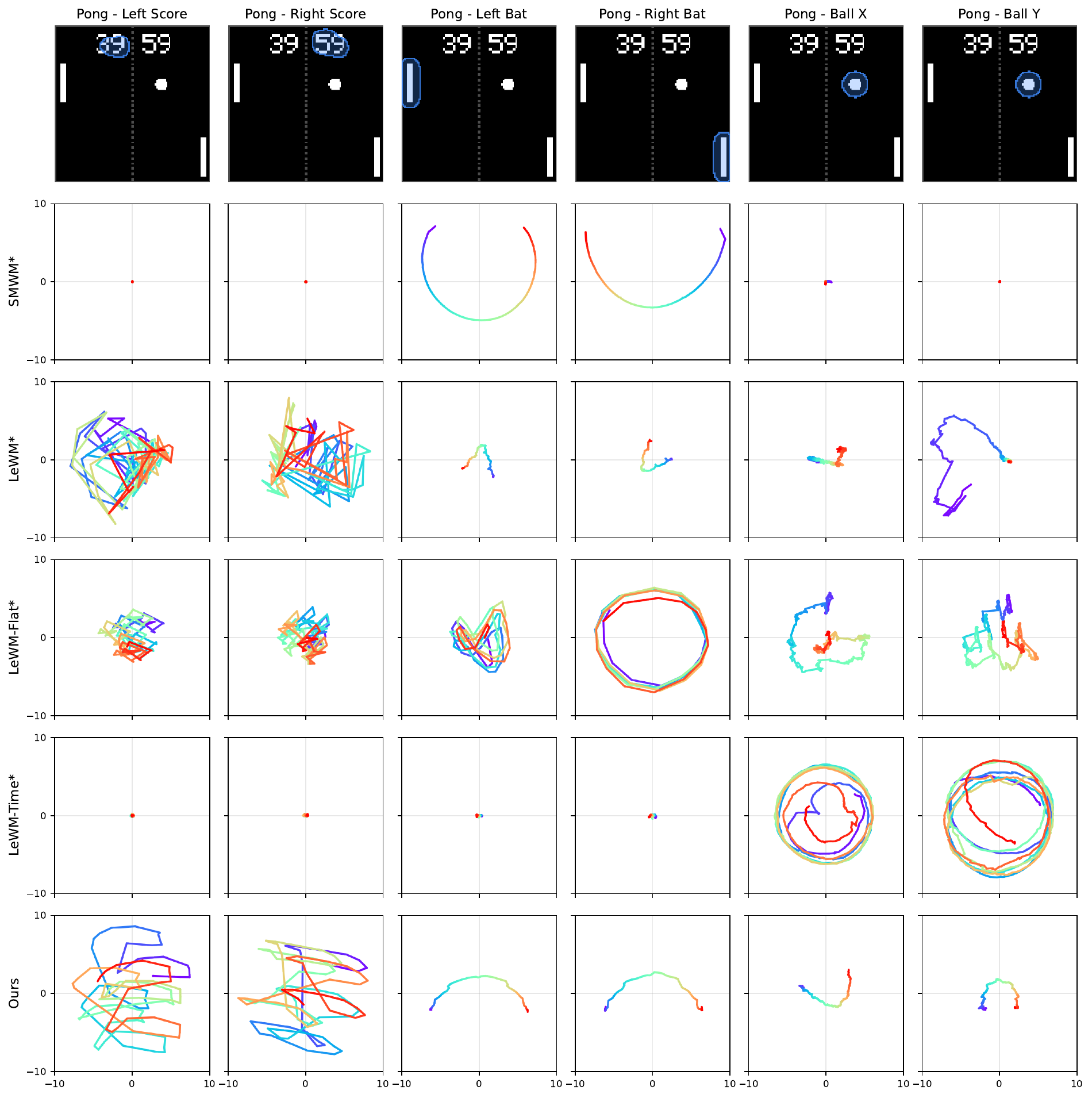}}
  \caption{\textbf{Per-Feature Latent Trajectories Visualization for Pong}.}
  \label{fig:appendix_games_latent_trajectory_pong}
\end{figure*}

\begin{figure*}
  \centering{\includegraphics[width=\textwidth]{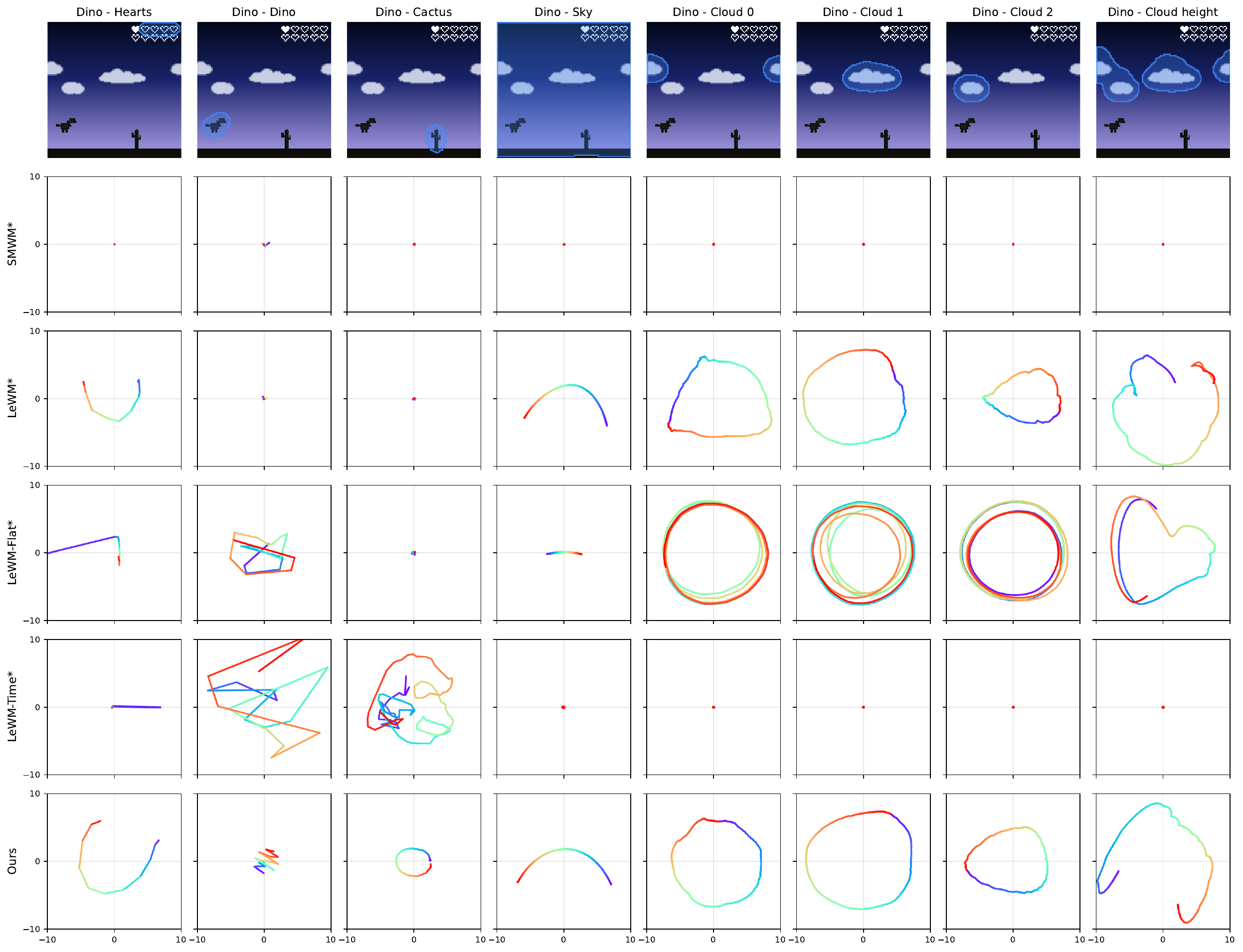}}
  \caption{\textbf{Per-Feature Latent Trajectories Visualization for Dino}.}
  \label{fig:appendix_games_latent_trajectory_dino}
\end{figure*}

\begin{figure*}
  \centering{\includegraphics[width=\textwidth]{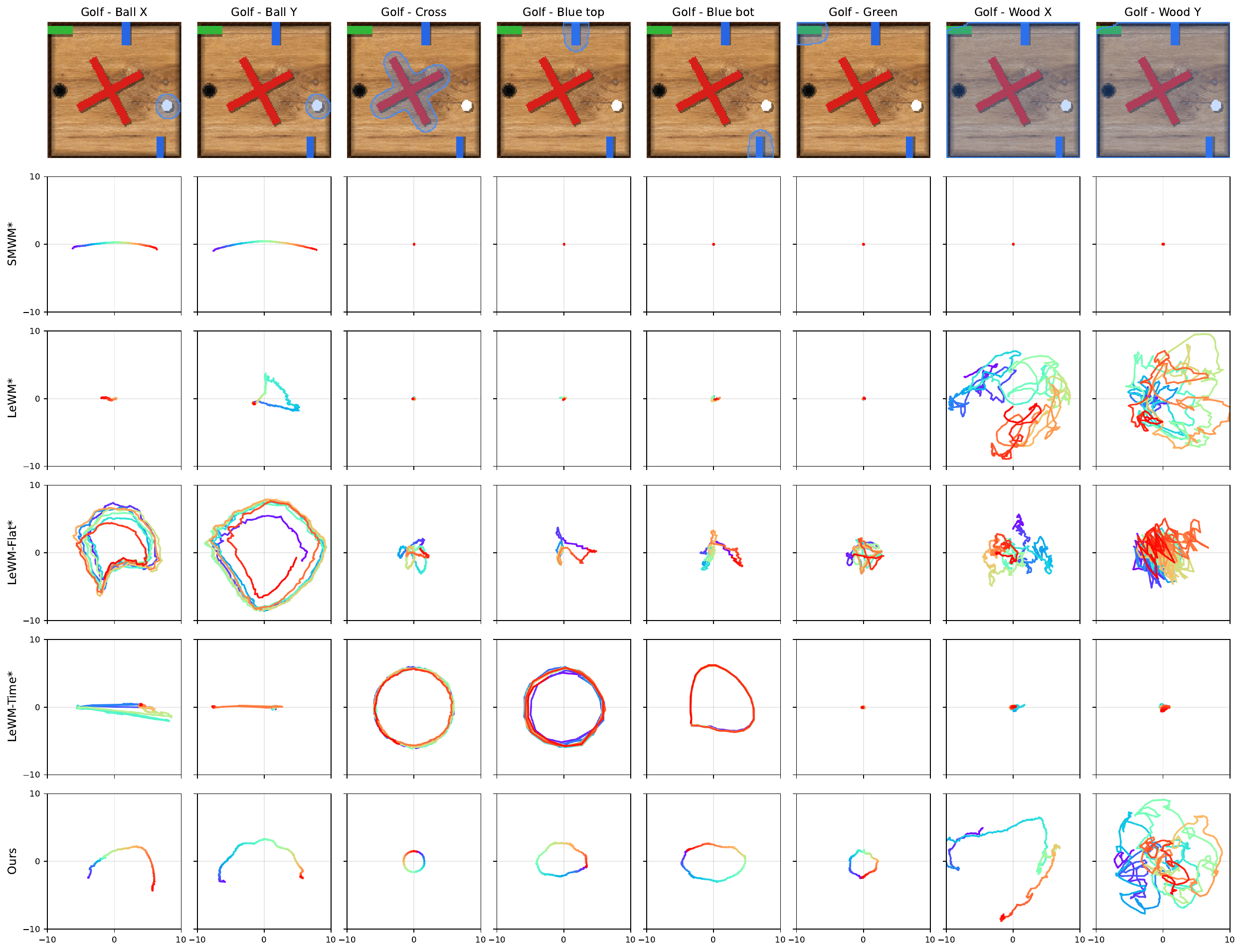}}
  \caption{\textbf{Per-Feature Latent Trajectories Visualization for Golf}.}
  \label{fig:appendix_games_latent_trajectory_golf}
\end{figure*}

\clearpage

\end{document}